%% file: 0_main.tex
\documentclass[10pt,twocolumn,letterpaper]{article}

\usepackage{iccv}
\usepackage{times}
\usepackage{epsfig}
\usepackage{graphicx}
\usepackage{amsmath}
\usepackage{amssymb}

\usepackage{caption} 
\usepackage[pagebackref=true,breaklinks=true,letterpaper=true,colorlinks,citecolor=blue,linkcolor=blue,bookmarks=false]{hyperref}

\iccvfinalcopy %

\def\iccvPaperID{6214} %

\ificcvfinal\fi

\input{macro}

\input{symbols}

\begin{document}

\title{Hybrid Neural Fusion for Full-frame Video Stabilization}

\author{
Yu-Lun Liu$^{1}$\quad
Wei-Sheng Lai$^{2}$\quad
Ming-Hsuan Yang$^{2,4,5}$\quad
Yung-Yu Chuang$^{1}$\quad
Jia-Bin Huang$^{3}$
\\
$^{1}$National Taiwan University \quad
$^{2}$Google \quad
$^{3}$Virginia Tech \quad
$^{4}$UC Merced \quad
$^{5}$Yonsei University\\
{\small\url{https://alex04072000.github.io/FuSta/}}
}

\ificcvfinal\thispagestyle{empty}\fi

\twocolumn[{%
\renewcommand\twocolumn[1][]{#1}%
\vspace{-5mm}
\maketitle
\vspace{-5mm}
\begin{center}
    \captionsetup{type=figure}
    \begin{minipage}[c]{0.02\textwidth}
    \rotatebox[origin = c]{90}{Input}
    \end{minipage}
    \begin{minipage}[c]{0.975\textwidth}
    \centering
        \includegraphics[height=0.1365\linewidth, width=0.2422\linewidth]{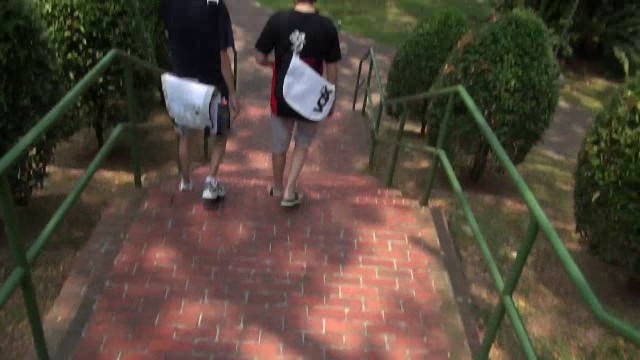} \hfill
        \includegraphics[height=0.1365\linewidth, width=0.2551\linewidth]{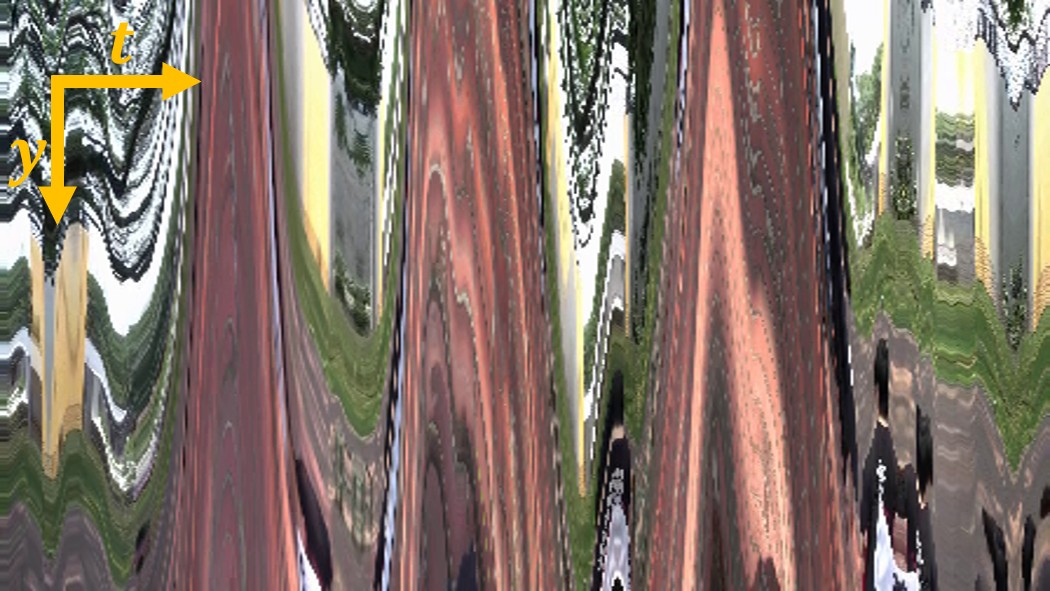} \hfill
        \includegraphics[height=0.1365\linewidth, width=0.2333\linewidth]{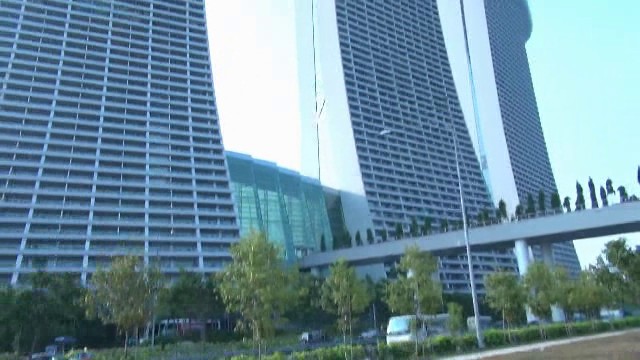} \hfill
        \includegraphics[height=0.1365\linewidth, width=0.2551\linewidth]{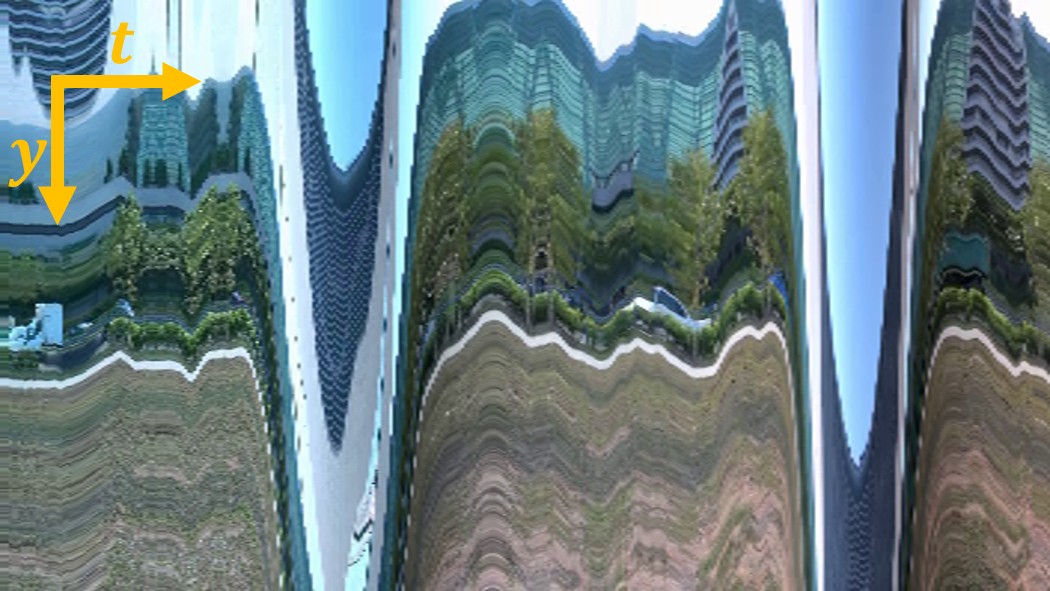}
    \end{minipage}
    \\
    \begin{minipage}[c]{0.02\textwidth}
    \rotatebox[origin = c]{90}{Our results}
    \end{minipage}
    \begin{minipage}[c]{0.975\textwidth}
    \centering
        \includegraphics[height=0.25725\linewidth, width=0.2422\linewidth]{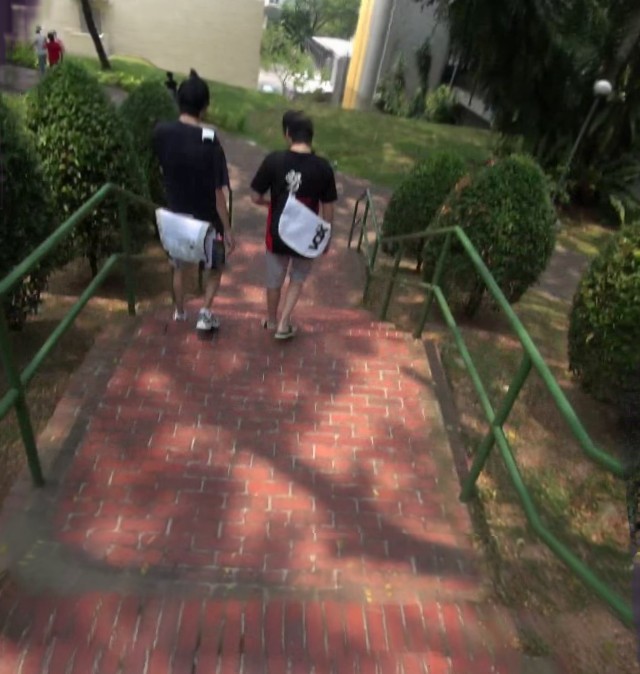}  \hfill
        \includegraphics[height=0.25725\linewidth, width=0.2551\linewidth]{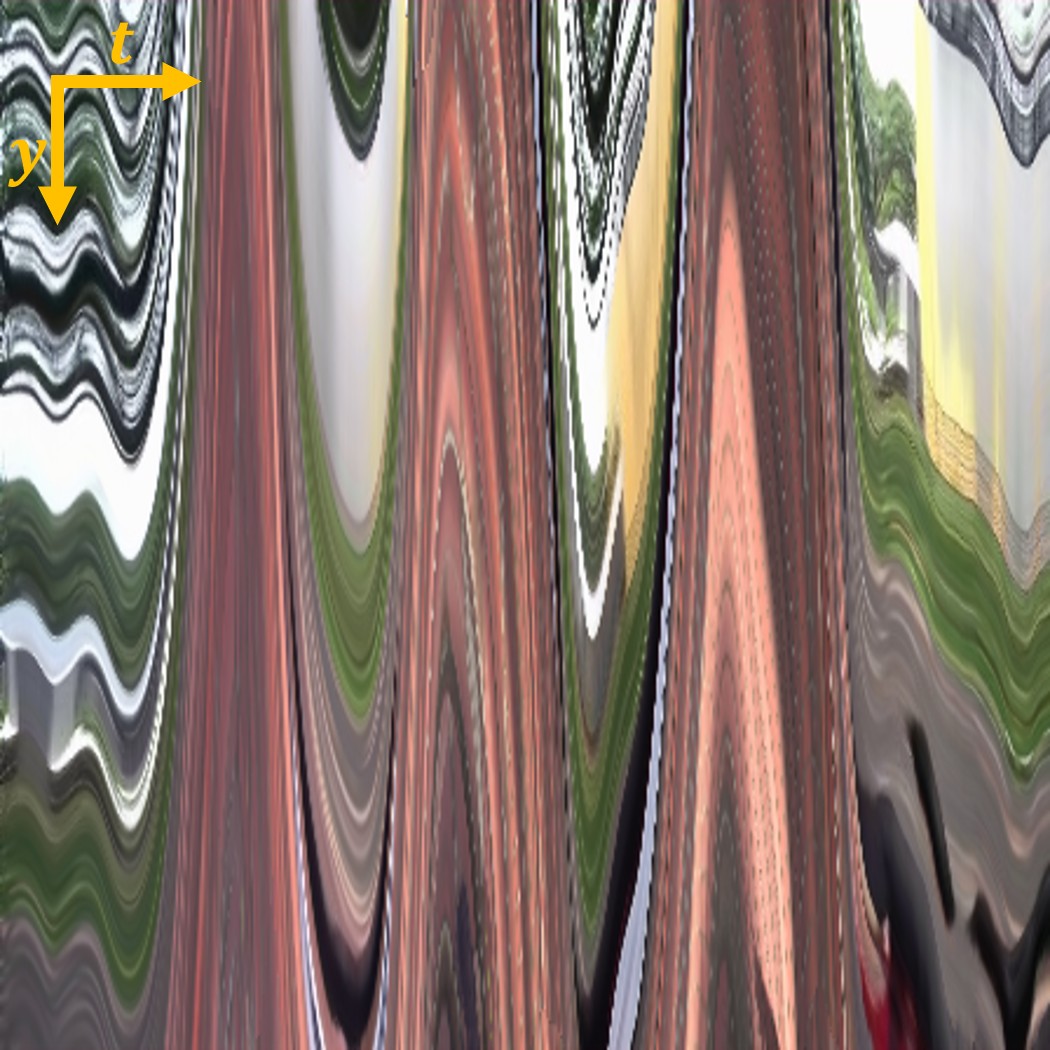}  \hfill
        \includegraphics[height=0.25725\linewidth, width=0.2333\linewidth]{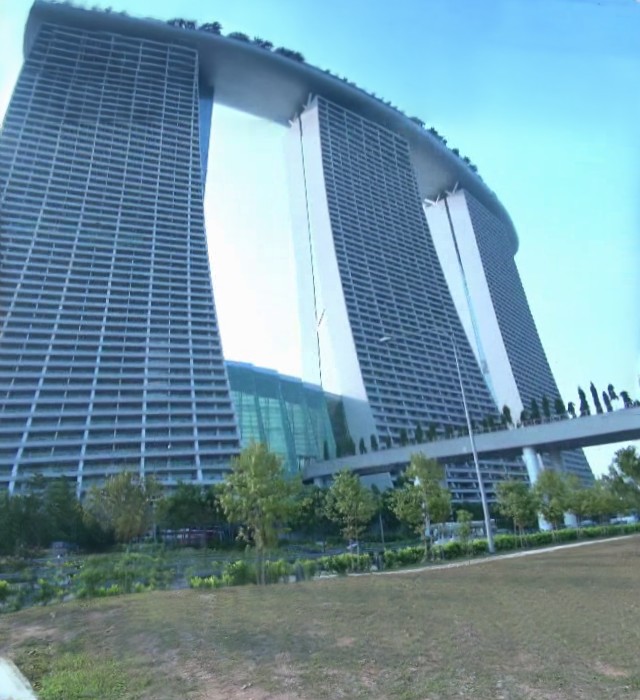}  \hfill           
        \includegraphics[height=0.25725\linewidth, width=0.2551\linewidth]{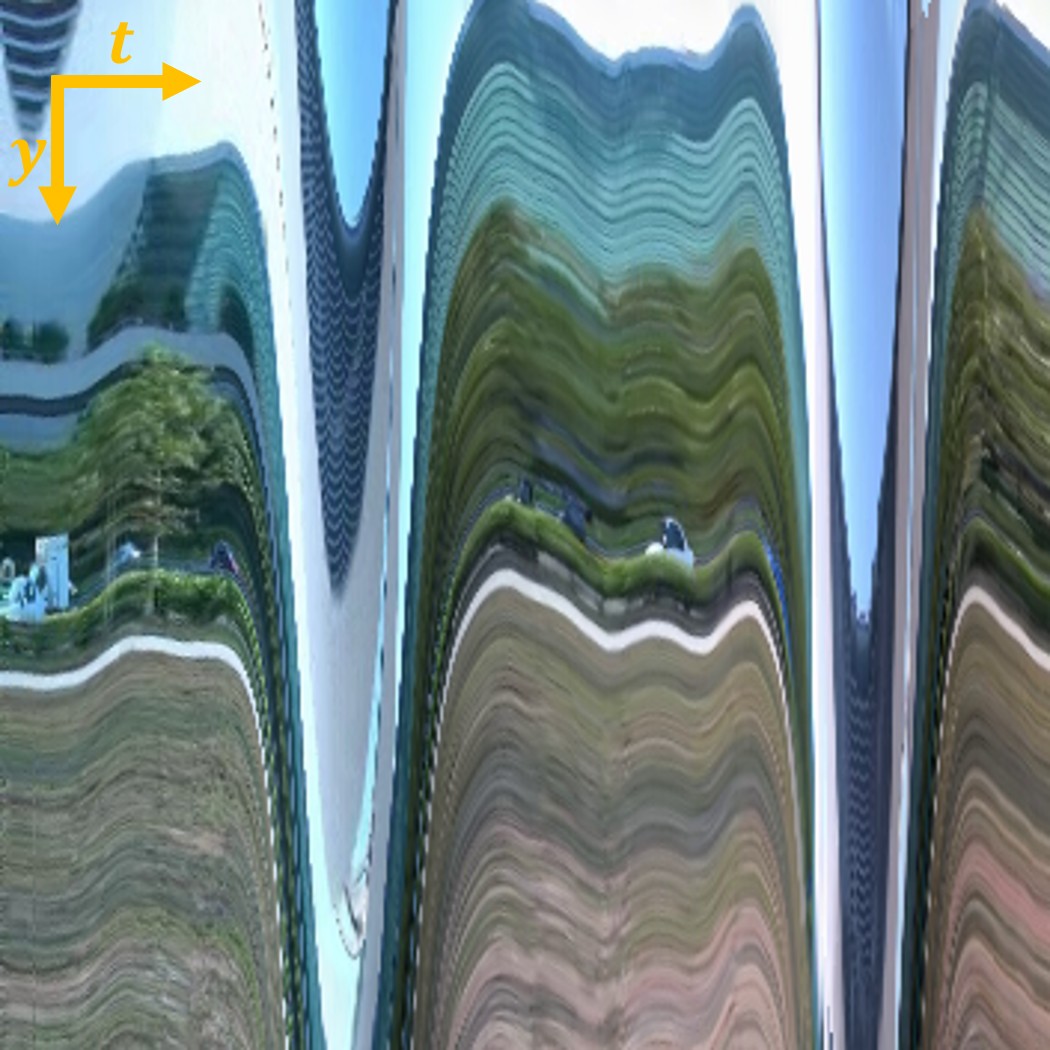}
    \end{minipage}
    \figmargin
    \caption{
        \textbf{Full-frame video stabilization of challenging videos.} 
        Our method takes a shaky input video (\emph{top}) and produces a stabilized and distortion-free video (\emph{bottom}), as indicated by less fluctuation in the $y\!-\!t$ epipolar plane image.
        Furthermore, by robustly fusing multiple neighboring frames, our results do not suffer from aggressive cropping of frame borders in the stabilized video and can even expand the field of view of the original video. 
        Our approach significantly outperforms representative state-of-the-art video stabilization algorithms on these challenging scenarios (see \figref{traditional_cropping}).
    } 
    \label{fig:teaser}
\end{center}
}]

\input{1_abstract}

\input{2_introduction}

\input{3_related_work}

\input{4_method}

\input{5_experiment}

\input{7_conclusion}

\vspace{1mm}
{
\small
\heading{Acknowledgments.} This work is supported in part by MOST 110-2221-E-002-124-MY3, 110-2634-F-002-026, and MediaTek Inc. We thank to National Center for High-performance Computing (NCHC) for providing computational and storage resources.
}

{\small
\bibliographystyle{ieee_fullname}
\bibliography{videoStab}
}

\end{document}

%% file: macro.tex
\usepackage{multirow}
\usepackage{caption} %
\usepackage{paralist} %
\usepackage{booktabs} %

\long\def\ignorethis#1{}

\usepackage{xcolor}
\usepackage{amsmath}
\usepackage{mathtools}%

\newcommand{\heading}[1]{
\vspace{1mm}\noindent\textbf{#1}}

\newcommand{\figref}[1]{Figure~\ref{fig:#1}}%
\newcommand{\tabref}[1]{Table~\ref{tab:#1}} %
\newcommand{\eqnref}[1]{Eq.~(\ref{eq:#1})}
\newcommand{\secref}[1]{Section~\ref{sec:#1}}

\definecolor{green}{rgb}{0.0, 0.5, 0.0}
\newcommand{\red}[1]{{\color{red}#1}}
\newcommand{\blue}[1]{{\color{blue}#1}}

\renewcommand{\paragraph}[1]{\vspace{1mm} \noindent\textit{#1}}

\newcommand{\figmargin}{\vspace{-2mm}}
\newcommand{\figcapmargin}{\vspace{-2mm}}

\newcommand{\ignore}[1]{}   %

\newcommand{\best}[1]{\red{\textbf{#1}}}
\newcommand{\second}[1]{\blue{#1}} %

\makeatletter
\newcommand{\providelength}[1]{%
  \@ifundefined{\expandafter\@gobble\string#1}
   {%
    \typeout{\string\providelength: making new length \string#1}%
    \newlength{#1}%
   }
   {%
    \sdaau@checkforlength{#1}%
   }%
}

\newcommand{\sdaau@checkforlength}[1]{%
  \edef\sdaau@temp{\expandafter\sdaau@getfive\meaning#1TTTTT$}%
  \ifx\sdaau@temp\sdaau@skipstring
    \typeout{\string\providelength: \string#1 already a length}%
  \else
    \@latex@error
      {\string#1 illegal in \string\providelength}
      {\string#1 is defined, but not with \string\newlength}%
  \fi
}
\def\sdaau@getfive#1#2#3#4#5#6${#1#2#3#4#5}
\edef\sdaau@skipstring{\string\skip}
\makeatother

%% file: symbols.tex
\def\frame{I}

\newcommand{\flow}[2]{F_{#1 \rightarrow #2}}

%% file: 1_abstract.tex
\begin{abstract}
Existing video stabilization methods often generate visible distortion or require aggressive cropping of frame boundaries, resulting in smaller field of views.
In this work, we present a frame synthesis algorithm to achieve full-frame video stabilization.
We first estimate dense warp fields from neighboring frames and then synthesize the stabilized frame by fusing the warped contents.
Our core technical novelty lies in the learning-based hybrid-space fusion that alleviates artifacts caused by optical flow inaccuracy and fast-moving objects. 
We validate the effectiveness of our method on the NUS, selfie, and DeepStab video datasets.
Extensive experiment results demonstrate the merits of our approach over prior video stabilization methods.
\end{abstract}

%% file: 2_introduction.tex
\section{Introduction}
\label{sec:intro}
Video stabilization has become increasingly important with the rapid growth of video content on the Internet platforms, such as YouTube, Vimeo, and Instagram.
Casually captured cellphone videos without a professional video stabilizer are often shaky and unpleasant to watch.
These videos pose significant challenges for video stabilization algorithms. 
For example, videos are often noisy due to small image sensors, particularly in low-light environments. 
Handheld captured videos may contain large camera shake/jitter, resulting in severe motion blur and wobble artifacts from a rolling shutter camera.
\vspace{-1pt}

\input{figs/fig_motivations}

\emph{Existing video stabilization methods} usually consist of three main components: 
1) motion estimation, 2) motion smoothing and 3) stable frame generation.
First, the motion estimation step involves estimating motion through 2D feature detection/tracking \cite{lee2009video,liu2011subspace,goldstein2012video,wang2013spatially}, dense flow~\cite{yu2019robust,yu2020learning}, or recovering camera motion and scene structures \cite{liu2009content,zhou2013plane,buehler2001non,smith2009light,liu2012video}.
Second, the motion smoothing step then removes the high-frequency jittering in the estimated motion and predicts the spatial transformations to stabilize each frame in the form of homography~\cite{matsushita2005full}, mixture of homography~\cite{liu2013bundled,grundmann2012calibration}, or per-pixel warp fields~\cite{liu2014steadyflow,yu2019robust,yu2020learning}.
Third, the stable frame generation step uses the predicted spatial transform to synthesize the stabilized video. 
The stabilized frames, however, often contain large missing regions at frame borders, particularly when videos with large camera motion.
This forces existing methods to apply aggressive cropping for maintaining a rectangular frame and therefore leads to a significantly \emph{zoomed-in} video with resolution loss (\figref{traditional_cropping}(a) and (b)).

\emph{Full-frame video stabilization methods} aim to address the above-discussed limitation and produce stabilized video with the same field of view (FoV).
One approach for full-frame video stabilization is to first compute the stabilized video (with missing pixels at the frame borders) and then apply flow-based video completion methods ~\cite{matsushita2005full,huang2016temporally,gao2020flow} to fill in missing contents.
Such two-stage methods may suffer from the inaccuracies in flow estimation and inpainting (e.g., in poorly textured regions, fluid motion, and motion blur). 
A recent learning-based method, DIFRINT~\cite{choi2020deep}, instead uses iterative frame interpolation to stabilize the video while maintaining the original FoV.
However, applying frame interpolation repeatedly leads to severe distortion and blur artifacts in challenging cases (\figref{traditional_cropping}(c)).

In this paper, we present a new algorithm that takes a shaky video and the estimated smooth motion fields for stabilization as inputs and produces a full-frame stable video.
The core idea of our method lies in fusing information from multiple neighboring frames in a robust manner. 
Instead of using color frames directly, we use a learned CNN representation to encode rich local appearance for each frame, fuse multiple aligned feature maps, and use a neural decoder network to render the final color frame.
We first explore multiple design choices for fusing and blending multiple aligned frames.
We then propose a hybrid fusion mechanism that leverages both feature-level and image-level fusion to alleviate the sensitivity to flow inaccuracy.
We further improve the visual quality of the synthesized results by learning to predict spatially varying blending weights, removing blurry input frames for sharp video generation, and transferring high-frequency details residual to the re-rendered, stabilized frames. 
To minimize regions where contents are unknown for all neighboring frames, we propose a path adjustment method for balancing the goals of smoothing camera motion and maximizing frame coverage.
Our method generates stabilized video with significantly fewer artifacts and distortions while retaining (or even expanding) the original FoV (\figref{teaser}).
We evaluate the proposed algorithm with state-of-the-art methods and commercial video stabilization software (Adobe Premiere Pro 2020 warp stabilizer). 
Extensive experiments show that our method performs favorably against existing methods on three public benchmark datasets \cite{liu2013bundled,yu2018selfie,wang2018deep}.
Our main contributions are:
\begin{compactitem}
\item We apply a neural fusion technique in the context of full-frame video stabilization to alleviate the issues of sensitivity to flow inaccuracy. 
\item We present a hybrid fusion method for fusing information from multiple stabilized frames at both feature- and image-level. 
We systematically validate various design choices through detailed ablation studies. 
\item We demonstrate favorable performance against representative video stabilization techniques on three public datasets. 
\end{compactitem}

%% file: figs/fig_motivations.tex
\providelength\liftup
\setlength\liftup{-3.5mm}

\begin{figure*}[t]
\centering
\includegraphics[width=0.245\textwidth, height=0.138\textwidth]{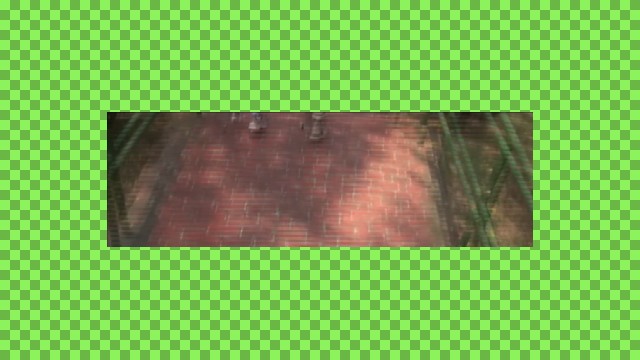}
\includegraphics[width=0.245\textwidth, height=0.138\textwidth]{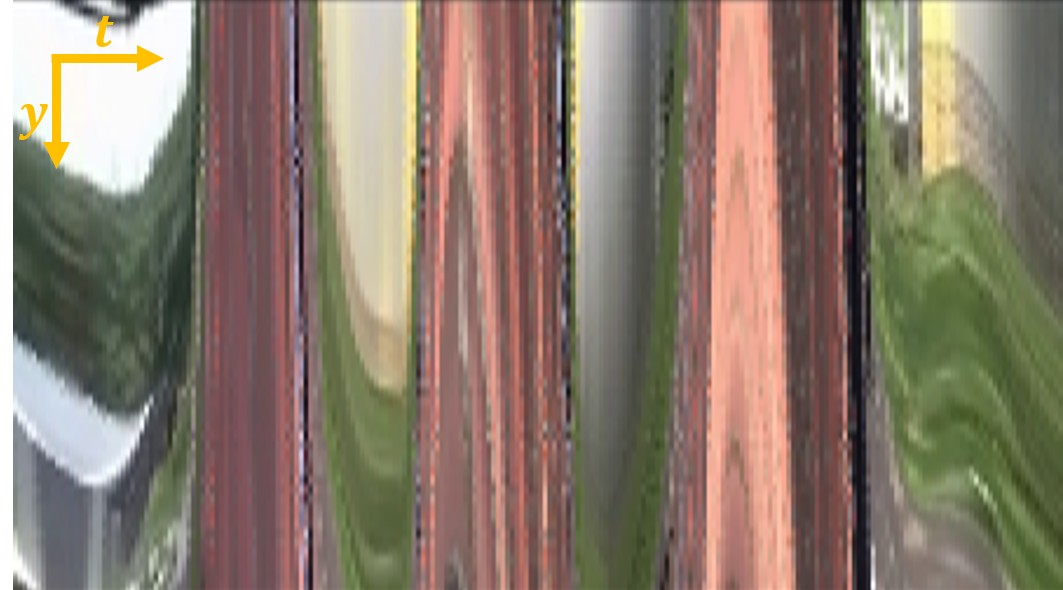} \hfill 
\includegraphics[width=0.245\textwidth, height=0.138\textwidth]{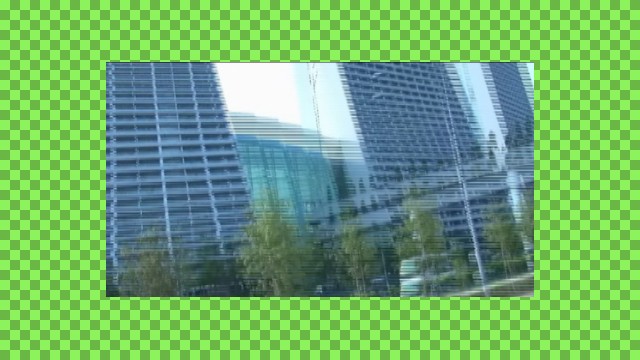}
\includegraphics[width=0.245\textwidth, height=0.138\textwidth]{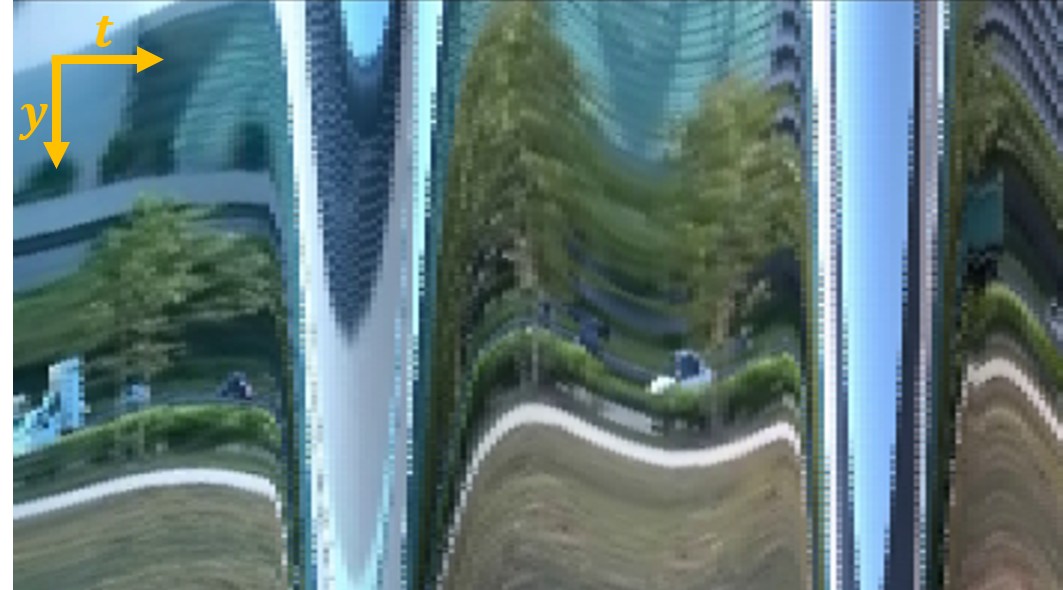}

\vspace{\liftup}

$\underbracket[1pt][2.0mm]{\hspace{\linewidth}}_{\substack{\vspace{-4.5mm}\\\colorbox{white}
{(a) Adobe Premiere Pro 2020 warp stabilizer}}}$
\includegraphics[width=0.245\textwidth, height=0.138\textwidth]{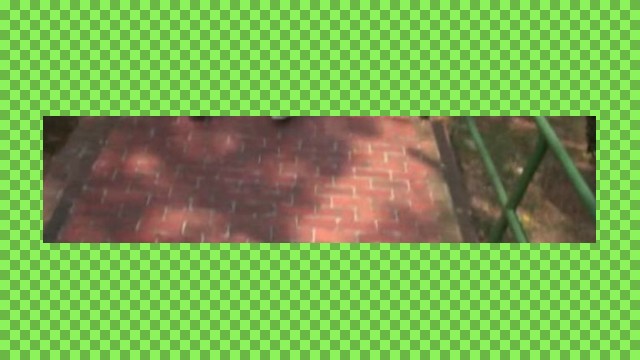} 
\includegraphics[width=0.245\textwidth, height=0.138\textwidth]{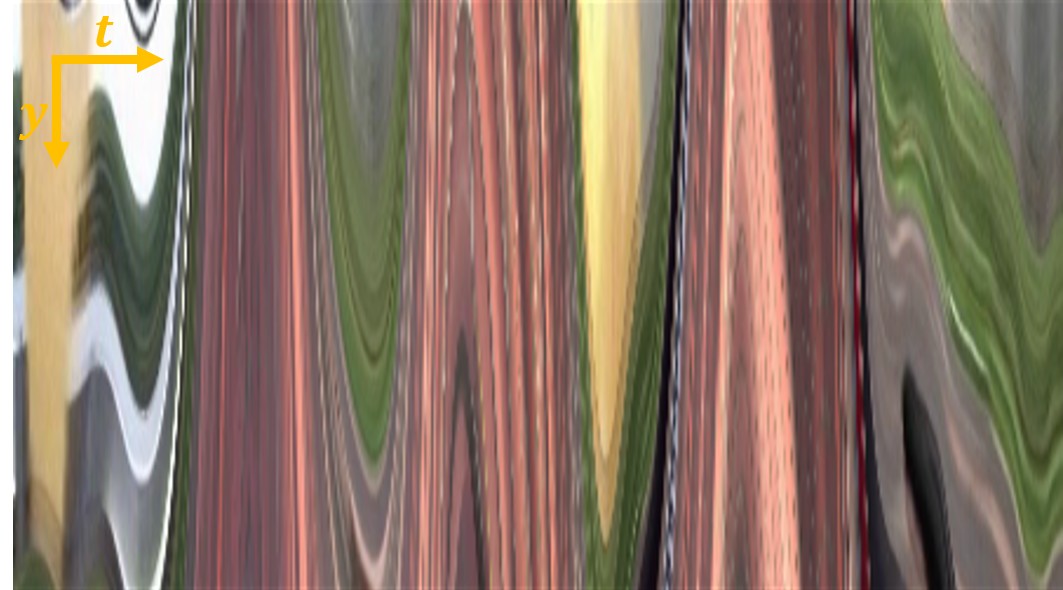} 
\hfill
\includegraphics[width=0.245\textwidth, height=0.138\textwidth]{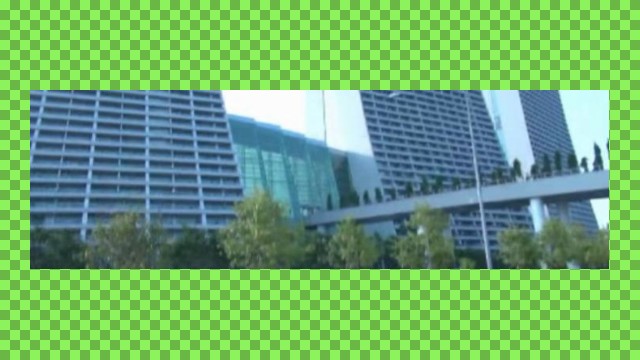}
\includegraphics[width=0.245\textwidth, height=0.138\textwidth]{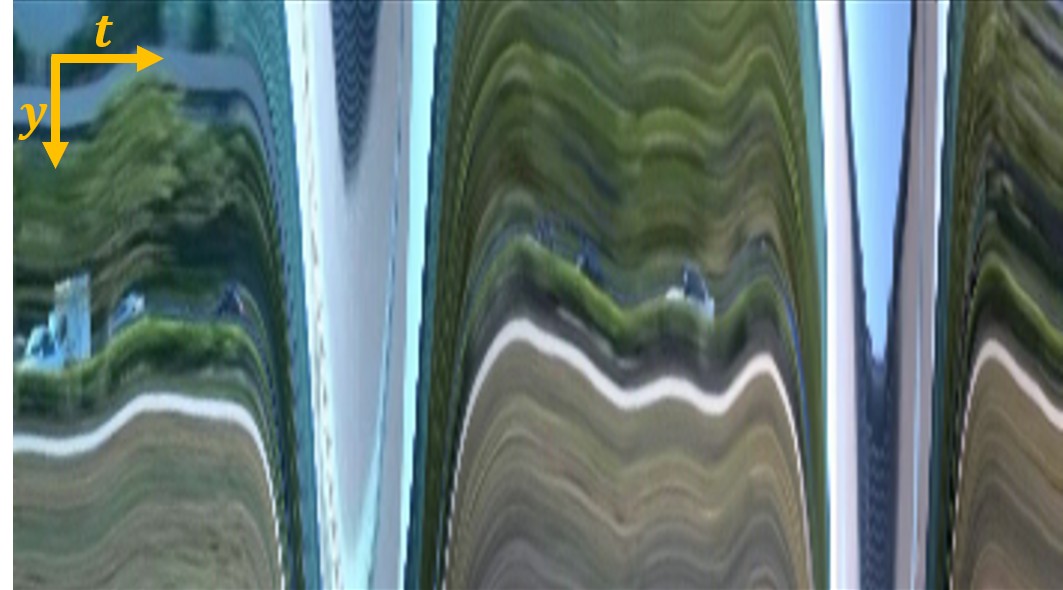} 

\vspace{\liftup}

$\underbracket[1pt][2.0mm]{\hspace{\linewidth}}_{\substack{\vspace{-4.5mm}\\\colorbox{white}
{(b) Yu and Ramamoorthi~\cite{yu2020learning}}}}$
\includegraphics[width=0.245\textwidth, height=0.138\textwidth]{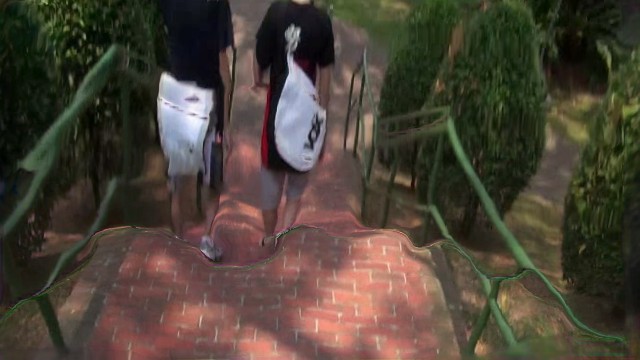} 
\includegraphics[width=0.245\textwidth, height=0.138\textwidth]{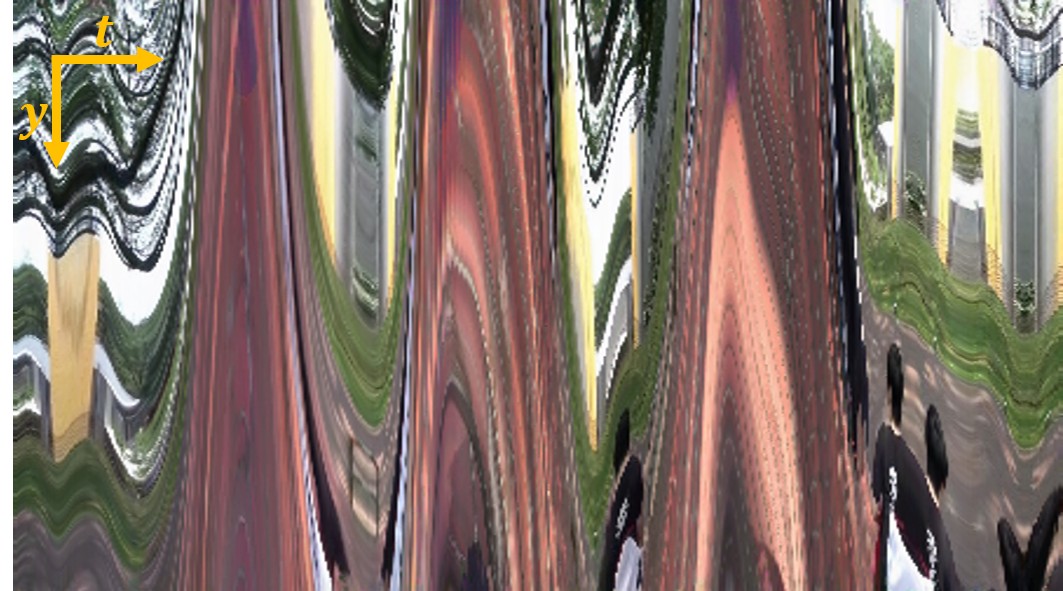} \hfill
\includegraphics[width=0.245\textwidth, height=0.138\textwidth]{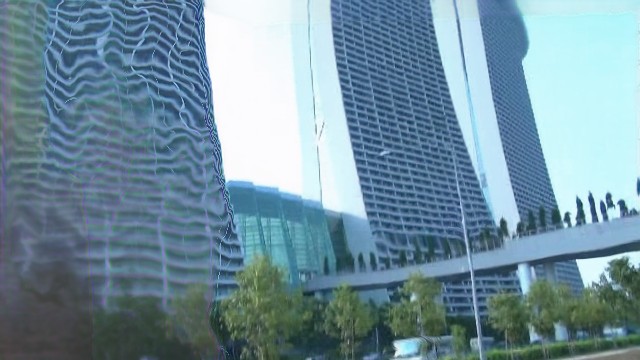}
\includegraphics[width=0.245\textwidth, height=0.138\textwidth]{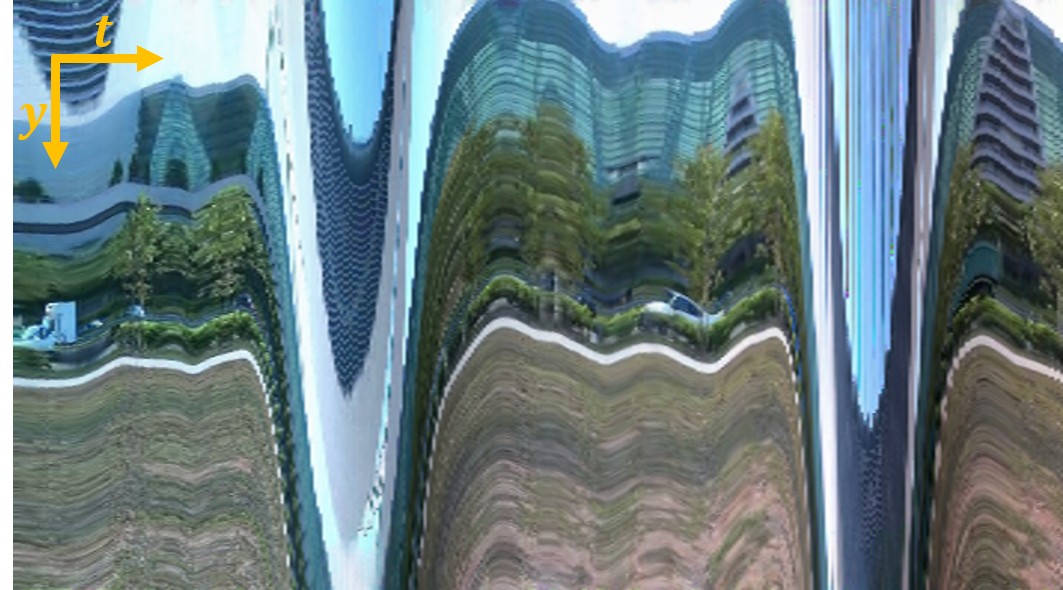}

\vspace{\liftup}

$\underbracket[1pt][2.0mm]{\hspace{\linewidth}}_{\substack{\vspace{-4.5mm}\\\colorbox{white}
{(c) DIFRINT~\cite{choi2020deep}}}}$

\vspace{-4mm}
\caption{
\textbf{Limitations of current state-of-the-art video stabilization techniques.}
(a) Current commercial video stabilization software (Adobe Premiere Pro 2020) fails to generate smooth videos in challenging scenarios of rapid camera shakes. 
(b) Yu and Ramamoorthi's method \cite{yu2020learning} produces a temporally smooth video. However, the warped (stabilized) video contains many missing pixels at frame borders and inevitably requires applying aggressive cropping (green checkerboard areas) to generate a rectangular video.
(c) The DIFRINT method~\cite{choi2020deep} achieves full-frame video stabilization by iteratively applying frame interpolation to generate in-between, stabilized frames.
However, interpolating between frames with large camera motion and moving occlusion is challenging. 
Their results are thus prone to severe artifacts. 
}
\vskip -8pt
\label{fig:traditional_cropping}
\end{figure*}

%% file: 3_related_work.tex
\section{Related work}
\label{sec:related}

\heading{Motion estimation and smoothing.}
Most video stabilization methods focus on estimating motion between frames and smoothing the motion.
They often estimate 2D motion using sparse feature detection/tracking and dense optical flow.
These methods differ in motion modeling, e.g., eigen-trajectories~\cite{liu2011subspace}, epipolar geometry~\cite{goldstein2012video}, warping grids~\cite{liu2013bundled}, or dense flow fields~\cite{yu2020learning}.
For motion smoothing, prior methods use low-pass filtering~\cite{liu2011subspace}, L1 optimization~\cite{grundmann2011auto}, and spatio-temporal optimization~\cite{wang2013spatially}.

In contrast to estimating 2D motion, several methods recover the camera motion and proxy scene geometry by leveraging Structure from Motion (SfM) algorithms.
These methods stabilize frames using 3D reconstruction and projection along with image-based rendering~\cite{kopf2014first} or content-preserving warps~\cite{buehler2001non,liu2009content,goldstein2012video}.
However, SfM algorithms are less effective in handling complex videos with severe motion blur and highly dynamic scenes~\cite{kopf2020robust}.
Specialized hardware such as depth cameras~\cite{liu2012video} or light field cameras~\cite{smith2009light} may be required for reliable pose estimation.

Deep learning-based approaches have recently been proposed to directly predict warping fields~\cite{xu2018deep,wang2018deep} or optical flows~\cite{yu2019robust,yu2020learning} for video stabilization.
In particular, methods with dense warp fields~\cite{liu2014steadyflow,yu2019robust,yu2020learning} offer greater flexibility for compensating motion jittering and implicitly handling rolling shutter effects than parametric warp fields~\cite{xu2018deep,wang2018deep}.

Our work builds upon existing 2D motion estimation/smoothing techniques for stabilization and focuses on synthesizing \emph{full-frame video} outputs. 
Specifically, we adopt the state-of-the-art flow-based stabilization method~\cite{yu2020learning} and use the estimated per-frame warped fields as inputs to our method.
Note that our method is \emph{agnostic} to the motion smoothing techniques. Other approaches such as parametric warps can also be applied.

\heading{Image fusion and composition.}
With the estimated and smoothed motion, the final step of video stabilization is to render the stabilized frames.
Most existing methods synthesize frames by directly warping each input frame to the stabilized location using smoothed warping grids~\cite{liu2013bundled, liu2011subspace} or flow fields predicted by CNNs~\cite{yu2019robust,yu2020learning}.
However, such approaches inevitably synthesize images with missing regions around frame boundaries.
To maintain a rectangle shape, existing methods often crop off the blank areas and generate output videos with a lower resolution and a smaller FOV than the input video.
To address this issue, full-frame video stabilization methods aim to stabilize videos without cropping.
These methods use neighboring frames to fill in the blank and produce full-frame results by 2D motion inpainting~\cite{matsushita2005full,huang2016temporally,gleicher2008re,gao2020flow}.
In contrast to existing motion inpainting methods that first generate stabilized frames then filling in missing pixels, our method leverage neural rendering to encode and fuse warped appearance features and learn to decode the fused feature map to the final color frames. 

Several recent methods can generate full-frame stabilized videos without explicit motion estimation. 
For example, the method in \cite{wang2018deep} train a CNN with collected unstable-stable pairs to directly synthesize stable frames.
However, direct synthesis of output frames without spatial transformations remains challenging.
Recently, the DIFRINT method~\cite{choi2020deep} generates full-frame stable videos by iteratively applying frame interpolation.
This method couples motion smoothing and frame rendering together. 
However, the repeated frame interpolation often introduces visible distortion and severe artifacts (see \figref{traditional_cropping}(c)).

\heading{View synthesis.}
View synthesis algorithms aim to render photorealistic images of novel viewpoints from a single image~\cite{niklaus20193d,wu2020unsupervised,wiles2020synsin,shih20203d,kopf2020one} or multiple posed images~\cite{levoy1996light,gortler1996lumigraph,chaurasia2013depth,penner2017soft,hedman2018deep,riegler2020free,riegler2020stable,sun2018multiview,choi2019extreme,saito2019pifu}. 
These methods mainly differ in the ways to map and fuse information, e.g., view interpolation~\cite{chen1993view,debevec1996modeling,seitz1996view}, 3D proxy geometry and mapping~\cite{buehler2001unstructured}, multi-plane images~\cite{zhou2018stereo,srinivasan2019pushing,huang2020semantic}, and CNN~\cite{hedman2018deep,riegler2020free,riegler2020stable}.
Our fusion network resembles the encoder-decoder network used for free view synthesis~\cite{riegler2020free}. 

A recent line of research focuses on rendering novel views for \emph{dynamic} scenes from a single video~\cite{xian2020space,tretschk2020non,li2020neural,gao2021dynamic} based on neural volume rendering~\cite{lombardi2019neural,mildenhall2020nerf}.
These methods can be used for full-frame video stabilization by rendering the dynamic video from a smooth camera trajectory.
While promising results have been shown, these methods require per-video training and precise camera pose estimates. 
In contrast, our frame synthesis method can be applied to a wider variety of videos without re-training and when accurate camera poses are difficult to obtain.

\heading{Neural rendering.}
A direct blending of multiple images in the image space may lead to glitching artifacts (visible seams).
Some recent methods train neural scene representations to synthesize novel views, such as NeRF~\cite{mildenhall2020nerf}, scene representation networks~\cite{sitzmann2019scene}, neural voxel grid~\cite{sitzmann2019deepvoxels,lombardi2019neural}, 3D Neural Point-Based Graphics~\cite{aliev2019neural}, and neural textures~\cite{thies2019deferred}.
However, these methods often require time-consuming per-scene training and do not handle dynamic scenes.
In contrast, our method does not require per-video finetuning.

%% file: 4_method.tex
\input{figs/fig_overview_two_columns}

\section{Full-frame video stabilization}
\label{sec:algorithm}
Let $\frame_{k^s}$ denote the \emph{source} frame in the real (unstabilized) camera space and $\frame_{k^t}$ the \emph{target} frame in the virtual (stabilized) camera space at a timestamp $k$.
Given an input video with $T$ frames $\{\frame_{k^s}\}_{k=1}^T$, our goal is to generate a video $\{\frame_{k^t}\}_{k=1}^T$ that is visually stable and maintains the same FOV as the input video without cropping.
Existing video stabilization methods often apply aggressive cropping to exclude any missing pixels due to frame warping, as shown in~\figref{traditional_cropping}.
In contrast, we utilize the information from neighboring frames to render stabilized frames with completed contents or even expanding the FOV of the input video.

Video stabilization methods typically consist of three stages: 1) motion estimation, 2) motion smoothing, and 3) frame warping/rendering.
Our method focuses on the third stage for rendering high-quality frames without any cropping.
Our proposed algorithm is thus agnostic to particular motion estimation/smooth techniques.
We assume that the warping field from the real camera space to the virtual camera space is available for each frame (e.g., from~\cite{yu2020learning}).

Given an input video, we first encode image features for each frame, warp the neighboring frames to the virtual camera space at the specific target timestamp, and then fuse the features to render a stabilized frame.
We describe the technical detail of each step in the following sections.

\subsection{Pre-processing}

\heading{Motion estimation and smoothing.}
Several motion estimation and smoothing methods have been developed~\cite{liu2011subspace, liu2013bundled, yu2020learning}.
This work uses the state-of-the-art method~\cite{yu2020learning} to obtain a backward dense warping field $\flow{k^t}{k^s}$ for each frame, where $k^s$ indicates the source input and ${k^t}$ denotes the target stabilized output.
These warping fields can be directly used to warp the input video.
However, the stabilized video often contains irregular boundaries and a large portion of missing pixels.
Therefore, the output video requires aggressive cropping and thus loses some content.

\heading{Optical flow estimation.}
To recover the missing pixels caused by warping, we need to project the corresponding pixels from nearby frames to the target stabilized frame.
For each key frame $I_{k^s}$ at time $k$, we compute the optical flows $\{\flow{n^s}{k^s}\}_{n\in\Omega_k}$ from neighboring frames to the key frame using RAFT~\cite{teed2020raft}, where $n$ indicates a neighboring frame and $\Omega_k$ denotes the set of neighboring frames for $I_{k^s}$. %

\subsection{Warping and fusion}
\label{sec:warping_and_fusion}

\heading{Warping.}
We warp the neighboring frames $\{\frame_{n^s}\} _{n\in\Omega_k}$ to align with the target frame $\frame_{k^t}$ in the virtual camera space.
Since we already have the warping field from the target frame to the keyframe $\flow{k^t}{k^s}$ (estimated from \cite{yu2020learning}) and the estimated optical flow from the keyframe to neighboring frames $\{ \flow{k^s}{n^s} \}_{n \in \Omega_k}$, we can then compute the warping field from the target frame to neighboring frames $\{ \flow{k^t}{n^s} \}_{n \in \Omega_k}$ by \emph{chaining} the flow vectors.
We can thus warp a neighboring frame $\frame_{n^s}$ to align with the target frame $\frame_{k^t}$ using backward warping~\cite{jaderberg2015spatial}.
Some pixels in the target frame are not visible in the neighboring frames due to occlusion/dis-occlusion or out-of-boundary.
Therefore, we compute visibility mask $\{M_{n^s}\}_{n\in\Omega_k}$ for each neighboring frame to indicate whether a pixel is valid (labeled as 1) in the source frame or not.
We use Sundaram~\etal's method~\cite{sundaram2010dense} to identify occluded pixels (labeled as 0).

\heading{Fusion space.}
With the aligned frames, we explore several fusion strategies. 
First, we can directly blend the warped color frames in the \emph{image space} to produce the output stabilized frame, as shown in~\figref{architecture_comparison}(a).
This image-space fusion approach is a commonly used technique in image stitching~\cite{agarwala2004interactive, szeliski2006image}, video extrapolation~\cite{lee2019video}, novel view synthesis~\cite{hedman2018deep}, and HDR reconstruction~\cite{kalantari2017deep}.
However, image-space fusion is prone to ghosting artifacts due to misalignment, or glitch artifacts due to inconsistent labeling between neighbor pixels.
Alternatively, one can also fuse the aligned frames in the \emph{feature space}, e.g., \cite{choi2019extreme, saito2019pifu}, as shown in \figref{architecture_comparison}(b).
Fusing in the high-dimensional feature spaces allows the model to be more robust to flow inaccuracy.
However, rendering the fused feature map using a neural image-translation decoder often leads to blurry outputs.

To combine the best worlds of both image-space and feature-space fusions, we propose a \emph{hybrid-space} fusion mechanism for video stabilization (\figref{architecture_comparison}(c)).
Similar to the feature-space fusion, we first extract high-dimensional features from each neighboring frame and warp the features using flow fields.
We then learn a CNN to predicting the blending weights that best fuse the features.
We concatenate the \emph{fused feature map} and the \emph{warped feature for each neighboring frame} to form the input for our image generator. 
The image generator learns to predict a target frame and a confidence map for each neighboring frame.
Finally, we adopt an image-space fusion to merge all the predicted target frames according to the predicted weights to generate the final stabilized frame.

The core difference between our hybrid-space fusion and feature-space fusion lies in the input to the image generator.
The image generator in~\figref{architecture_comparison}(b) takes \emph{only} the fused feature as input to predict the output frame.
The fused feature map already contains mixed information from multiple frames.
The image generator may thus have difficulty in synthesizing sharp image contents. 
In contrast, our image generator in \figref{architecture_comparison}(c) takes the fused feature map as guidance to reconstruct the target frame from the warped feature.
We empirically find that this improves the sharpness of the output frame while avoiding ghosting and glitching artifacts, as shown in the supplementary material.

\heading{Fusion function.}
We explore a \emph{learning-based} fusion method using deep CNNs. 
Specifically, we train a CNN to predicts a blending weight $\omega_{n^s}^{k^t}$ for each neighboring frame using the encoded features, visibility masks, and the flow error (\figref{fusion_choices}):
\begin{equation} \label{eq:CNN_fusion}
    f_{\text{CNN}}^{k^t} = \sum_{n\in \Omega_k} f_{n^s}^{k^t} \underbrace{\sigma(G_\theta(f_{n^s}^{k^t}, M_{n^s}^{k^t}, f_{k^s}^{k^t}, M_{k^s}^{k^t}, e_{n^s}^{k^t}))}_{\omega_{n^s}^{k^t}},
\end{equation}
where $G_\theta$ is the CNN, $\sigma(\cdot)$ is a softmax activation, $f_{n^s}^{k^t}$ and $M_{n^s}^{k^t}$ are the encoded feature map and warping mask of frame $n$, respectively. 
The superscript $k^t$ indicates that the encoded feature and warping mask are warped to the target stable frame $k$.
The forward-backward flow consistency error $e_{n^s}$ is calculated by:
\begin{equation} \label{eq:flow_error}
    e_{n^s}(\textbf{p}) =  \| {F_{{k^s} \rightarrow {n^s}} (\textbf{p}) + F_{{n^s} \rightarrow {k^s}} (\textbf{p} + F_{{k^s} \rightarrow {n^s}})}\|_2,
\end{equation}
where $\textbf{p}$ denotes the pixel coordinate in $F_{{k^s} \rightarrow {n^s}}$.
The error $e_{n^s}^{k^t}$ in~\eqnref{CNN_fusion} is calculated by warping the flow consistency error $e_{n^s}$ to the target frame $k$.
All the flow consistency errors are calculated from the input unstabilized frames.

After fusing the feature, we concatenate the fused feature with the warped feature and warping mask of each frame as the input to the image generator.
The image generator then predicts the output color frame and confidence map for each frame:
\begin{equation} \label{eq:image_decoder}
    \{I_{n^s}^{k^t}, C_{n^s}^{k^t}\} = G_\phi (f_{n^s}^{k^t}, M_{n^s}^{k^t}, f_{\text{CNN}}^{k^t}),
\end{equation}
where $G_\phi$ denotes the image generator, $I_{n^s}^{k^t}$ and $C_{n^s}^{k^t}$ represent the predicted frame and confidence map of frame $n$ in the virtual camera space at time $k$, respectively.
Finally, the output stabilized frame $I_{k^t}$ is generated by a weighted sum using these predicted frames and confidence maps:
\begin{equation} \label{eq:CNN_fusion_image_space}
    I_{k^t} = \sum_{n\in\Omega_k} {I_{n^s}^{k^t} C_{n^s}^{k^t}}.
\end{equation}

\input{figs/fig_learned_fusion}

\begin{table*}[t]
\centering
\footnotesize
\caption{\textbf{Quantitative evaluation of fusion functions and fusion spaces on the test set of~\cite{su2017deep}.} We highlight \best{the best} and \second{the second best} in each column.
}
\label{tab:ablation_fusion_function}
\renewcommand{\tabcolsep}{5pt} %
\begin{tabular}{l|cccccccccccc}
\hline
 & & \multicolumn{3}{c}{(a) Image-space fusion} & & \multicolumn{3}{c}{(b) Feature-space fusion} & & \multicolumn{3}{c}{(c) Hybrid-space fusion} \\ \cline{3-5} \cline{7-9} \cline{11-13}
 & &   LPIPS $\downarrow$ &  SSIM $\uparrow$ &  PSNR $\uparrow$  & & LPIPS $\downarrow$ & SSIM $\uparrow$ & PSNR $\uparrow$  & & LPIPS $\downarrow$ & SSIM $\uparrow$ & PSNR $\uparrow$ \\
\hline
Multi-band blending%
& & 0.105 & \second{0.926} & 26.150 & & - & - & - & & - & - & - \\
Graph-cut%
& & 0.105 & \best{0.928} & 26.190 & & - & - & - & & - & - & - \\
Mean & & 0.123 & 0.899 & 24.618 & & 0.108 & 0.878 & 26.028 & & 0.099 & 0.898 & 27.013\\
Gaussian & & \best{0.099} & 0.920 & 25.310 & & 0.095 & 0.874 & 26.344 & & 0.097 & 0.899 & 27.080\\
Argmax & & 0.105 & \best{0.928} & \second{26.200} & & \second{0.093} & \second{0.892} & \second{26.891} & & \second{0.087} & 0.906 & \second{27.519} \\
Flow error-weighted & & 0.114 & 0.901 & 25.363 & & 0.096 & 0.885 & 26.176 & & 0.095 & \second{0.911} & 27.371\\
CNN-based (Ours) & & \second{0.101} & 0.895 & \best{26.692} & & \best{0.092} & \best{0.902} & \best{27.187} & & \best{0.073} & \best{0.914} & \best{27.868}\\
\hline
\end{tabular}%

\end{table*}

\subsection{Path adjustment}
\label{sec:path_adjustment}
When stabilizing a long video with large motion, some pixels around the frame boundary in the target frames may not be visible in \emph{any} of the neighboring frames.
For such cases, the network has to ``hallucinate'' new contents, often resulting in blurry predictions and unwanted visual artifacts.

To mitigate this problem, we propose to adjust the flow fields by global translations in each frame to increase the coverage of valid regions in the entire video. 
More specifically, for each target frame $k$, we aim to find a global translation $\textbf{x}_{k^t}$ to adjust the flow fields of its neighbor frames as $\flow{k^t}{n^s} + \textbf{x}_{k^t}$, where $n\in\Omega_k$.
Let the valid pixel mask $M^{k^t}_\text{valid}(\textbf{x}_{k^t})$ denote the \emph{union} of the warping masks after the adjustment.
We find the translations by optimizing the following energy function:
\begin{equation} \label{eq:alpha_expansion}
    \mathop{\arg\min}_{\textbf{x}_{k^t}} \sum_{k^t} {(1-M^{k^t}_\text{valid}(\textbf{x}_{k^t}))} +
    \lambda_\text{s} \sum_{k^t, q^t} {\left \| \textbf{x}_{k^t} - \textbf{x}_{q^t} \right \|^2_2},
\end{equation}
where $q \in \{k\pm 1\}$ indicates the neighbor frame of frame $k$. 
Here, the first term is a data term that aims to maximize the \emph{coverage} of the valid mask. 
The second term is a smoothness term that penalizes large translation adjustments between nearby frames. 
The weight $\lambda_\text{s}$ is a hyper-parameter that balances between the data and smoothness terms. 
We set $\lambda_\text{s}=100$ in all the experiments and solve~\eqnref{alpha_expansion} using the coarse-to-fine alpha-expansion approach~\cite{boykov2004experimental}.

\subsection{Training details}
\heading{Loss functions.}
Our loss functions include the L1 and VGG perceptual losses:
\begin{equation} \label{eq:loss}
    \mathcal{L} = \|| I_{k^t} - \hat{I}_{k^t} \||_1 + \sum_{l} \lambda_l \|| \psi_l(I_{k^t}) - \psi_l(\hat{I}_{k^t}) \||_1,
\end{equation}
where $\hat{I}_{k^t}$ denotes the ground-truth target frame and $\psi_l$ is intermediate feature extracted from a pre-trained VGG-19 network~\cite{simonyan2014very}.

\heading{Training data.}
Our model requires a pair of unstabilized and stabilized videos for training.
However, it is difficult to obtain such a real video pair.
Therefore, we apply random motion to a stable video sequence to synthesize the input shaky video.
Specifically, we sample a short sequence of 7 frames from the training set of~\cite{su2017deep} and randomly crop the frames to generate the input \emph{unstable} video.
We then apply another random cropping on the center frame as the ground-truth of the target \emph{stabilized} frame.

%% file: figs/fig_overview_two_columns.tex
\begin{figure*}[t]
\centering
\renewcommand{\tabcolsep}{1pt} %
\renewcommand{\arraystretch}{1} %
\begin{tabular}{ccc}
\includegraphics[width=0.275\linewidth]{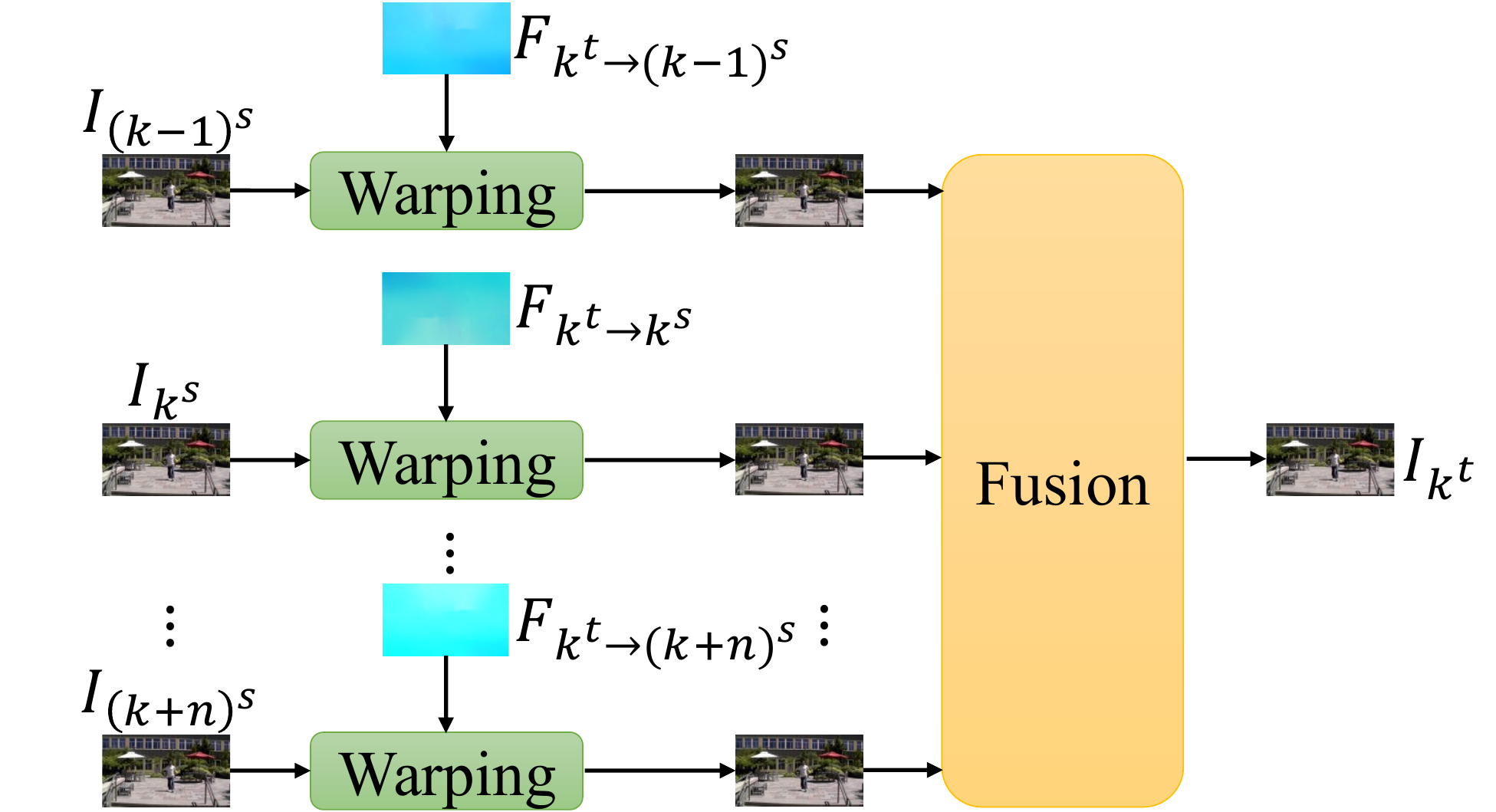} &\includegraphics[width=0.315\linewidth]{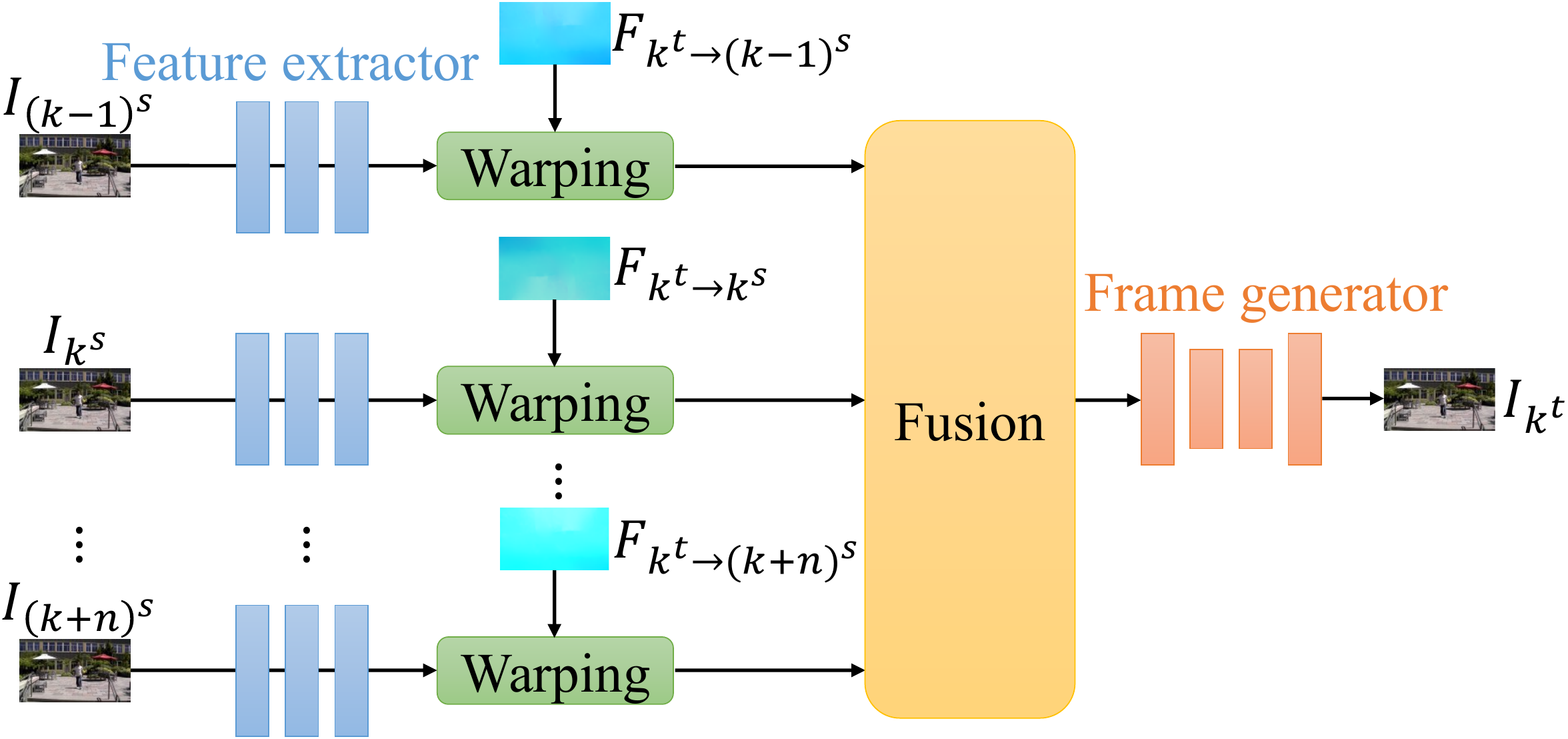} &\includegraphics[width=0.395\linewidth]{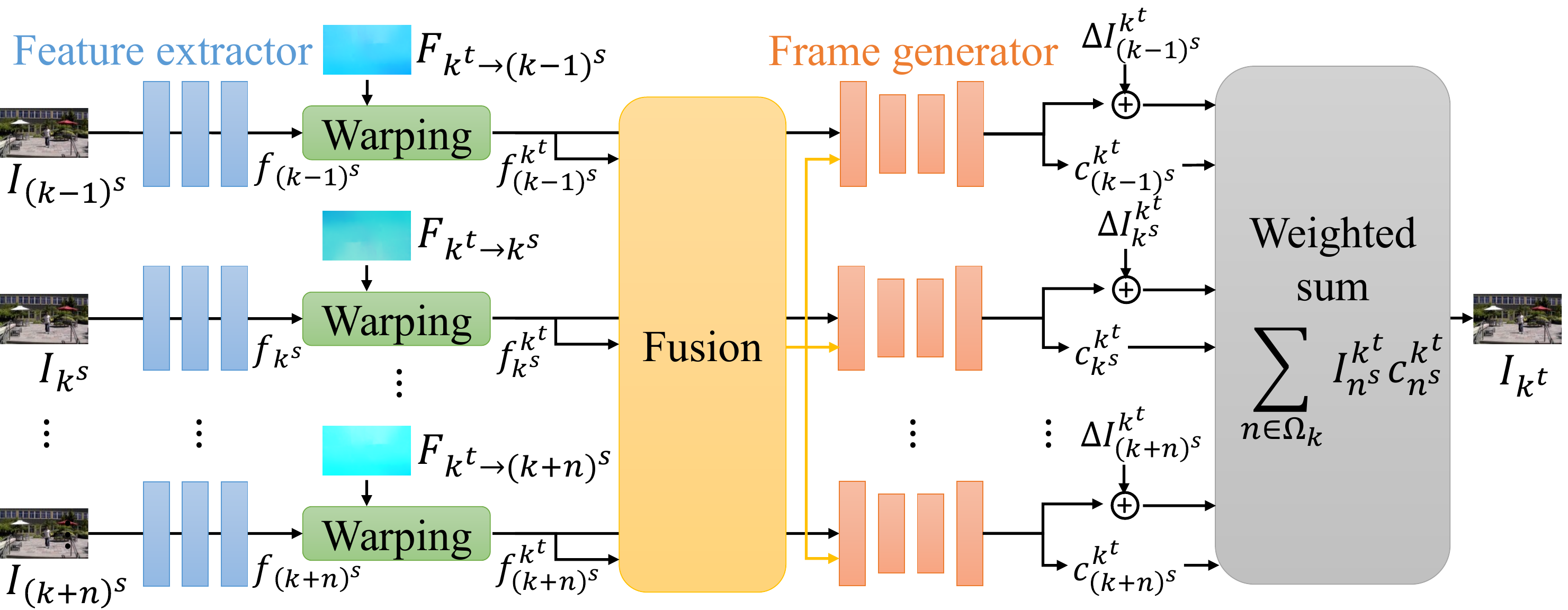} \\

(a) Image (late fusion) & (b) Feature (early fusion) & (c) Hybrid fusion (proposed) \\

\end{tabular}
\figcapmargin
\caption{
\textbf{Design choices for fusing multiple frames.}
To synthesize a full-frame stabilized video, we need to \emph{align} and \emph{fuse} the contents from multiple neighboring frames in the input shaky video.
(a) Conventional panorama image stitching (or in general image-based rendering) methods often fuse the warped (stabilized) images in the \emph{image level}. 
Fusing in image-level works well when the alignment is accurate, but may generate blending artifacts (e.g., visible seams) when flow estimates are not reliable.
(b) One can also encode the images as abstract CNN features, perform the fusion in the feature-space, and learn a decoder to convert the fused feature to output frames. 
Such approaches are more robust to flow inaccuracy but often produce overly blurred images.
(c) Our proposed hybrid fusion combines the advantages of both strategies. 
We first extract abstract image features (\eqnref{CNN_fusion}).
We then fuse the warped features from multiple frames. 
For each source frame, we take the fused feature map together with the individual warped features and decode it to the output frames and the associated confidence maps.
Finally, we produce the final output frame by using the weighted average of the generated images as in ~\eqnref{CNN_fusion_image_space}.
}
\label{fig:architecture_comparison}
\end{figure*}

%% file: figs/fig_learned_fusion.tex
\begin{figure}[t]
\centering
\includegraphics[width=\linewidth]{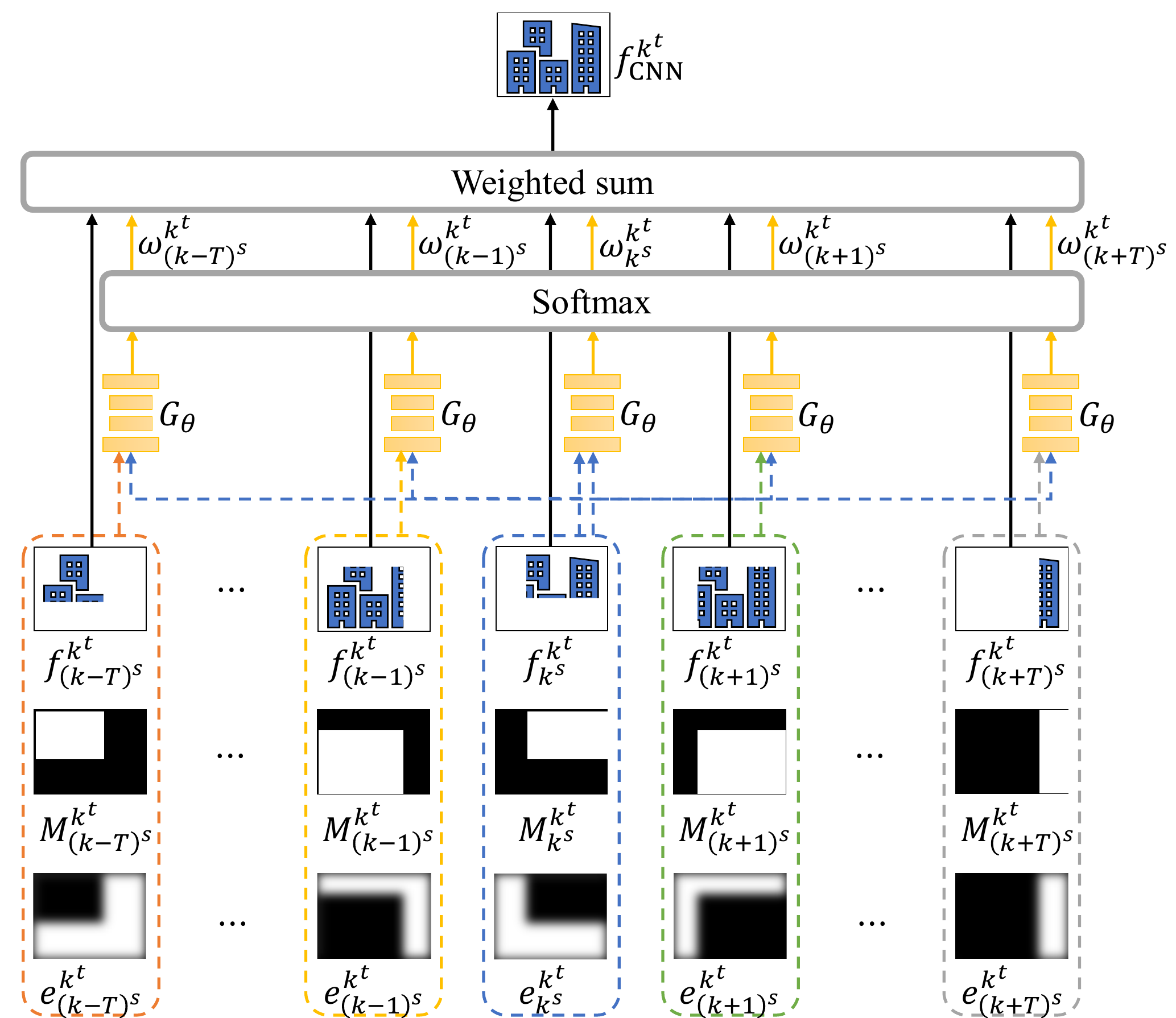}
\caption{
\textbf{Learning-based fusion.}
Given the warped features $\{f_{n^s}^{k^t}\}_{n\in\Omega_k}$ (or warped images for image-space fusion), warping masks $\{M_{n^s}^{k^t}\}_{n\in\Omega_k}$, and the flow error maps $\{e_{n^s}^{k^t}\}_{n\in\Omega_k}$, 
we first concatenate feature, warping mask, and flow erro maps for each frame (shown as dotted blocks).
We then use a CNN to predict the blending weights $\omega_{n^s}^{k^t}$ for each neighbor frame. 
Using the predicted weights, we compute the fused feature by weighted averaging the individual warped features $\{f_{n^s}^{k^t}\}_{n\in\Omega_k}$.
}
\label{fig:fusion_choices}
\end{figure}

%% file: 5_experiment.tex
\section{Experimental results}
\label{sec:experiments}

We start with validating various design choices of our approach (\secref{ablation}).
Next, we present quantitative comparison against representative state-of-the-art video stabilization algorithms (\secref{quantitative}) and visual results (\secref{visual}). Finally, we evaluate our fusion method on view synthesis (\secref{viewsynthesis}).
In the supplementary material, we further present 1) the definitions of the fusion functions, evaluation metrics, and the technical details of the network architecture, 2) the ablation studies on path adjustment, residual detail transfer, 3) comparisons with video completion method~\cite{gao2020flow} and learning-based blending~\cite{hedman2018deep}, 4) the user study, 5) the limitations of our approach, and 6) stabilized videos by the evaluated methods.

\subsection{Ablation study}
\label{sec:ablation}
We analyze the contribution of each design choice, including the fusion function design and fusion mechanism.
We generate 940 test videos using the same method as our training data generation, where each video contains seven input frames and one target frame.

\heading{Fusion function.}
We explore the following differentiable functions for fusion:
1) \emph{Mean fusion}, 
2) \emph{Gaussian-weighted fusion}, 
3) \emph{Argmax fusion}, 
4) \emph{Flow error-weighted fusion}, and 
5) the proposed \emph{CNN-based fusion function}.
We train the proposed model using image-space fusion, feature-space fusion, and hybrid-space fusion.
For image-space fusion, we also include two conventional fusion methods: multi-band blending~\cite{brown2003recognising} and graph-cut~\cite{agarwala2004interactive}.

\tabref{ablation_fusion_function} shows the quantitative results of different fusion methods and fusion spaces.
None of the fusion methods dominates the results for image-space fusion, where the argmax and CNN-based fusions perform slightly better than other alternatives.
For both feature-space and hybrid-space fusion, the proposed CNN-based fusion
shows advantages over other approaches.

\begin{table*}[t]
\small
\centering
\footnotesize
\caption{\textbf{Quantitative evaluations with the state-of-the-art methods on the NUS dataset~\cite{liu2013bundled}, the selfie dataset~\cite{yu2018selfie}, and the DeepStab dataset~\cite{wang2018deep}.} 
We evaluate the following metrics: \emph{Cropping Ratio} (C), \emph{Distortion Value} (D), \emph{Stability Score} (S), and \emph{Accumulated Optical Flow} (A). 
\best{Red} text indicates the best and \second{blue} text indicates the second-best performing method.
}
\label{tab:Metric_NUS}
\renewcommand{\tabcolsep}{5pt} %
\begin{tabular}{lcccccccccccccc}
\hline
& \multicolumn{4}{c}{NUS dataset~\cite{liu2013bundled}} & & \multicolumn{4}{c}{Selfie dataset~\cite{yu2018selfie}} & & \multicolumn{4}{c}{DeepStab dataset~\cite{wang2018deep}} \\ \cline{2-5} \cline{7-10} \cline{12-15}
 & C $\uparrow$ & D $\uparrow$ & S $\uparrow$ & A $\downarrow$ & & C $\uparrow$ & D $\uparrow$ & S $\uparrow$ & A $\downarrow$ & & C $\uparrow$ & D $\uparrow$ & S $\uparrow$ & A $\downarrow$ \\
\hline
Bundle~\cite{liu2013bundled} & 0.84 & \second{0.93} & \best{0.89} & \second{0.78} & & 0.68 & 0.82 & 0.85 & 0.84 & & 0.76 & 0.91 & \second{0.84} & \best{0.56} \\
L1Stabilizer~\cite{grundmann2011auto} & 0.74 & 0.92 & \best{0.89} & 0.88 & & 0.75 & \best{0.92} & 0.85 & 0.84 & & 0.74 & 0.92 & \best{0.85} & 0.70 \\
StabNet~\cite{wang2018deep} (online method) & 0.66 & 0.88 & 0.82 & 1.02 & & 0.70 & 0.78 & 0.83 & 0.83 & & 0.65 & 0.86 & 0.80 & 0.80 \\
DIFRINT~\cite{choi2020deep} & \best{1.00} & \best{0.96} & 0.83 & 0.87 & & \best{1.00} & 0.87 & 0.85 & \second{0.72} & & \best{1.00} & \second{0.94} & 0.78 & 0.78 \\
Yu and Ramamoorthi~\cite{yu2020learning} & \second{0.86} & 0.91 & 0.85 & 0.88 & & 0.78 & 0.79 & \second{0.87} & 0.77 & & \second{0.82} & 0.92 & 0.81 & 0.72 \\
Yu and Ramamoorthi~\cite{yu2018selfie}  & - & - & - & - & & \second{0.85} & \second{0.91} & \best{0.88} & 0.76 & & - & - & - & - \\
Adobe Premiere Pro 2020 warp stabilizer & 0.74 & 0.82 & \second{0.87} & 0.84 & & 0.71 & 0.80 & 0.84 & 0.79 & & 0.73 & 0.87 & 0.83 & 0.78 \\
Ours & \best{1.00} & \best{0.96} & 0.85 & \best{0.77} & & \best{1.00} & 0.87 & \second{0.87} & \best{0.64} & & \best{1.00} & \best{0.96} & 0.81 & \second{0.64} \\
\hline
\end{tabular}
\end{table*}

\input{figs/fig_visual_comparison}

\heading{Fusion space.}
Next, we compare different fusion levels using our CNN-based fusion.
The last row of \tabref{ablation_fusion_function} shows that the proposed hybrid-space fusion achieves the best results compared to image-space fusion and feature-space fusion.
The synthesized frame from the image-space fusion looks sharp but often contains visible glitching artifacts due to the discontinuity of different frames and inaccurate motion estimation.
The results of the feature-space fusion are smooth but overly blurred.
Finally, our hybrid-space fusion takes advantage of both above methods, generating a sharp and artifact-free frame.

\subsection{Quantitative evaluation}
\label{sec:quantitative}
We evaluate the proposed method with state-of-the-art video stabilization algorithms, including L1Stabilizer~\cite{grundmann2011auto}, Bundle~\cite{liu2013bundled}, StabNet~\cite{wang2018deep}, DIFRINT~\cite{choi2020deep}, and Yu and Ramamoorthi~\cite{yu2020learning}, and the warp stabilizer in Adobe Premiere Pro CC 2020.
StabNet is the only online method for performance evaluation.
It is included because it provides the DeepStab dataset.
We obtain the results of the compared methods from the videos released by the authors or generated from the publicly available official implementation with default parameters or pre-trained models.

\heading{Datasets.}
We evaluate the above methods on the NUS dataset~\cite{liu2013bundled}, the selfie dataset~\cite{yu2018selfie}, and the DeepStab dataset~\cite{wang2018deep}.
The NUS dataset consists of 144 video sequences, which are classified into six categories: Simple, Quick rotation, Zooming, Parallax, Crowd, and Running.
The selfie dataset includes 33 video clips with frontal faces and severe jittering.
The DeepStab dataset includes 61 videos with forward, pan, spin, and complex movements.

\heading{Metrics.}
We use the following metrics to evaluate the performance of video stabilization, which are widely used in prior work~\cite{liu2013bundled, wang2018deep, choi2020deep, yu2019robust}: 
\textbf{1)} \emph{Cropping ratio}: measures the remaining frame area after cropping off the undefined pixels due to motion compensation. 
\textbf{2)} \emph{Distortion value}: measures the anisotropic scaling of the homography between the input and output frames. 
\textbf{3)} \emph{Stability score}: measures the stability and smoothness of the stabilized video. 
\textbf{4)} \emph{Accumulated optical flow}: accumulates the optical flow over the entire stabilized video. 
Note that these metrics are used to evaluate the performance of video stabilization, while in~\secref{ablation} we use PSNR, SSIM~\cite{wang2004image}, and LPIPS~\cite{zhang2018perceptual} to evaluate the quality of synthesized frames.

\heading{Results on the NUS dataset.}
We show the average scores on the NUS dataset~\cite{liu2013bundled} on the left side of~\tabref{Metric_NUS}.
Both DIFRINT\cite{choi2020deep} and our method are full-frame methods and thus have an average cropping ratio of 1.
Note that the distortion metric measures the global distortion by fitting a homography between the input and stabilized frames. Therefore, it is not suitable to measure local distortion.
Although DIFRINT~\cite{choi2020deep} obtains the highest distortion score, its results contain visible local or color distortion due to iterative frame interpolation, as shown in~\figref{visual_comparison} and the supplementary material.
Adobe Premiere Pro 2020 warp stabilizer obtains the highest stability score at the cost of a low cropping ratio (0.74).
Our results achieve the best distortion score and accumulated optical flow, a good stability score, and an average cropping ratio of 1 (no cropping for all the videos), demonstrating our advantages over the state-of-the-art approaches in the NUS dataset.

\begin{table}[t]
\footnotesize
\centering
\caption{\textbf{Using different flow smoothing methods in our framework.}
Our method improves three metrics on the Selfie dataset~\cite{yu2018selfie} when integrating with different flow smoothing methods.}
\label{tab:Metric_bundle_flow}
\renewcommand{\tabcolsep}{5pt} %
\begin{tabular}{l|cccc}
\hline
 & C $\uparrow$ & D $\uparrow$ & S $\uparrow$ & A $\downarrow$ \\
\hline
Bundle~\cite{liu2013bundled} & 0.68 & 0.82 & \textbf{0.85} & 0.84 \\
Ours + Bundle~\cite{liu2013bundled} & \textbf{1.00} & \textbf{0.84} & \textbf{0.85} & \textbf{0.70} \\
\hline
Yu and Ramamoorthi~\cite{yu2020learning} & 0.78 & 0.79 & \textbf{0.87} & 0.77 \\
Ours + Yu and Ramamoorthi~\cite{yu2020learning} & \textbf{1.00} & \textbf{0.87} & \textbf{0.87} & \textbf{0.64} \\
\hline
\end{tabular}
\end{table}

\heading{Results on the Selfie dataset.}
The middle of~\tabref{Metric_NUS} shows the average scores on the selfie dataset~\cite{yu2018selfie}.
Note that the videos in the dataset often contain large camera motions and are more challenging to stabilize.
Similar to the NUS dataset, our method achieves the best cropping ratio and accumulated flow.
Our distortion and stability scores are comparable to Yu and Ramamoorthi's method~\cite{yu2018selfie}, which is specially designed to stabilize selfie videos.

\heading{Results on the DeepStab dataset.}
We show the average scores on the DeepStab dataset~\cite{wang2018deep} on the right side of~\tabref{Metric_NUS}. Our method achieves the best distortion, second-best accumulated flow, and good stability without cropping.

\heading{Integrating with different flow smoothing methods.}
The proposed method can be easily integrated with existing flow smoothing methods to generate a full-frame stabilized video. \tabref{Metric_bundle_flow} shows that the proposed method not only obtains full-frame results but also improves all metrics when combining with other smoothing methods \cite{liu2013bundled} or \cite{yu2020learning}.

\subsection{Visual comparison}
\label{sec:visual}
We show the stabilized frame of our method and state-of-the-art approaches from the Selfie dataset in~\figref{visual_comparison}.
Most of the methods~\cite{grundmann2011auto,wang2018deep,liu2013bundled,yu2020learning} suffer from a large amount of cropping, as indicated by the green checkerboard regions.
The aspect ratios of the output frames are also changed.
To keep the same aspect ratio as the input video, excessive cropping is required.
DIFRINT~\cite{choi2020deep} generates full-frame results.
However, its results often contain visible artifacts due to frame interpolation.
In contrast, our method generates full-frame stabilized videos with fewer visual artifacts.

\subsection{View Synthesis}
\label{sec:viewsynthesis}
Although we develop our fusion method for video stabilization, it is not limited to stabilization, and we show that it can be integrated into other multi-image fusion tasks such as video completion and FOV expansion in the supplementary. 
The proposed hybrid fusion method can also be used for view synthesis.
We compare our approach to state-of-the-art novel view synthesis methods on the Tanks and Temples dataset~\cite{Knapitsch2017}.
\tabref{FVS} and~\figref{FVS} show that our method performs favorably against other compared view synthesis methods quantitatively and qualitatively.
Our method achieves the best results among all the methods in all evaluated metrics.
Visually, our method generates images with fewer artifacts and more details than FVS~\cite{Riegler2020FVS}.

\begin{table}[t]
\centering
\small
\renewcommand{\tabcolsep}{1pt} %
\renewcommand{\arraystretch}{1} %
\caption{\textbf{Quantitative comparison of view synthesis.}}
\label{tab:FVS}
\resizebox{\columnwidth}{!}{%
\begin{tabular}{lrrrrrrrrrrrrrrr}
\toprule
& \multicolumn{3}{c}{Truck} & & \multicolumn{3}{c}{Train} & & \multicolumn{3}{c}{M60} & & \multicolumn{3}{c}{Playground} \\ \cline{2-4} \cline{6-8} \cline{10-12} \cline{14-16}
 &  LPIPS &  SSIM &  PSNR & &  LPIPS &  SSIM &  PSNR & &  LPIPS &  SSIM &  PSNR & &  LPIPS &  SSIM &  PSNR \\ 
\midrule
EVS~\cite{choi2019extreme} & 0.41 & 0.563 & 14.99 & & 0.64 & 0.454 & 11.81 & & 0.62 & 0.473 & 9.66 & & 0.39 & 0.610 & 16.34 \\
LLFF~\cite{mildenhall2019llff} & 0.61 & 0.432 & 10.66 & & 0.70 & 0.356 & 8.88 & & 0.69 & 0.427 & 8.98 & & 0.56 & 0.517 & 13.27 \\
NeRF~\cite{mildenhall2020nerf} & 0.61 & 0.690 & 19.47 & & 0.74 & 0.532 & 13.16 & & 0.62 & 0.691 & 15.99 & & 0.54 & 0.734 & 21.16 \\
NPBG~\cite{aliev2019neural} & 0.22 & 0.822 & 20.32 & & \best{0.25} & \best{0.801} & \second{18.08} & & 0.36 & 0.716 & 12.35 & & 0.17 & \best{0.876} & \second{23.03} \\
FVS~\cite{Riegler2020FVS} & \second{0.15} & \second{0.875} & \second{22.21} & & 0.32 & 0.759 & 17.46 & & \second{0.30} & \second{0.785} & \second{17.13} & & \second{0.16} & 0.849 & 22.32 \\
Ours & \best{0.12} & \best{0.882} & \best{23.25} & & \second{0.26} & \second{0.778} & \best{18.76} & & \best{0.27} & \best{0.788} & \best{17.43} & & \best{0.14} & \second{0.854} & \best{23.20} \\
\bottomrule
\end{tabular}
}
\end{table}

\begin{figure}[t]
\centering
\small
\renewcommand{\tabcolsep}{1pt} %
\renewcommand{\arraystretch}{1} %
\begin{tabular}{ccc}
\includegraphics[width=0.15\textwidth]{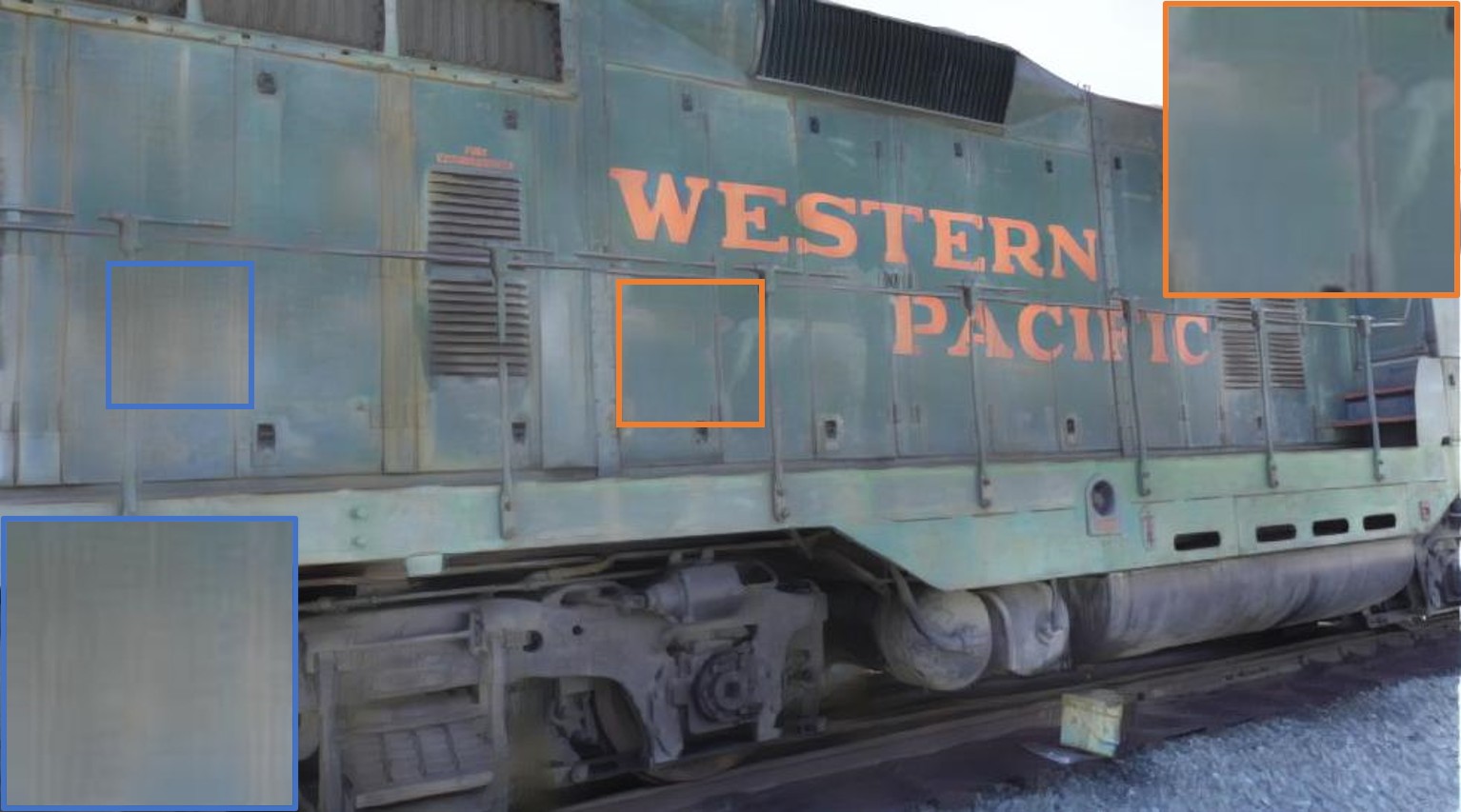} & 
\includegraphics[width=0.15\textwidth]{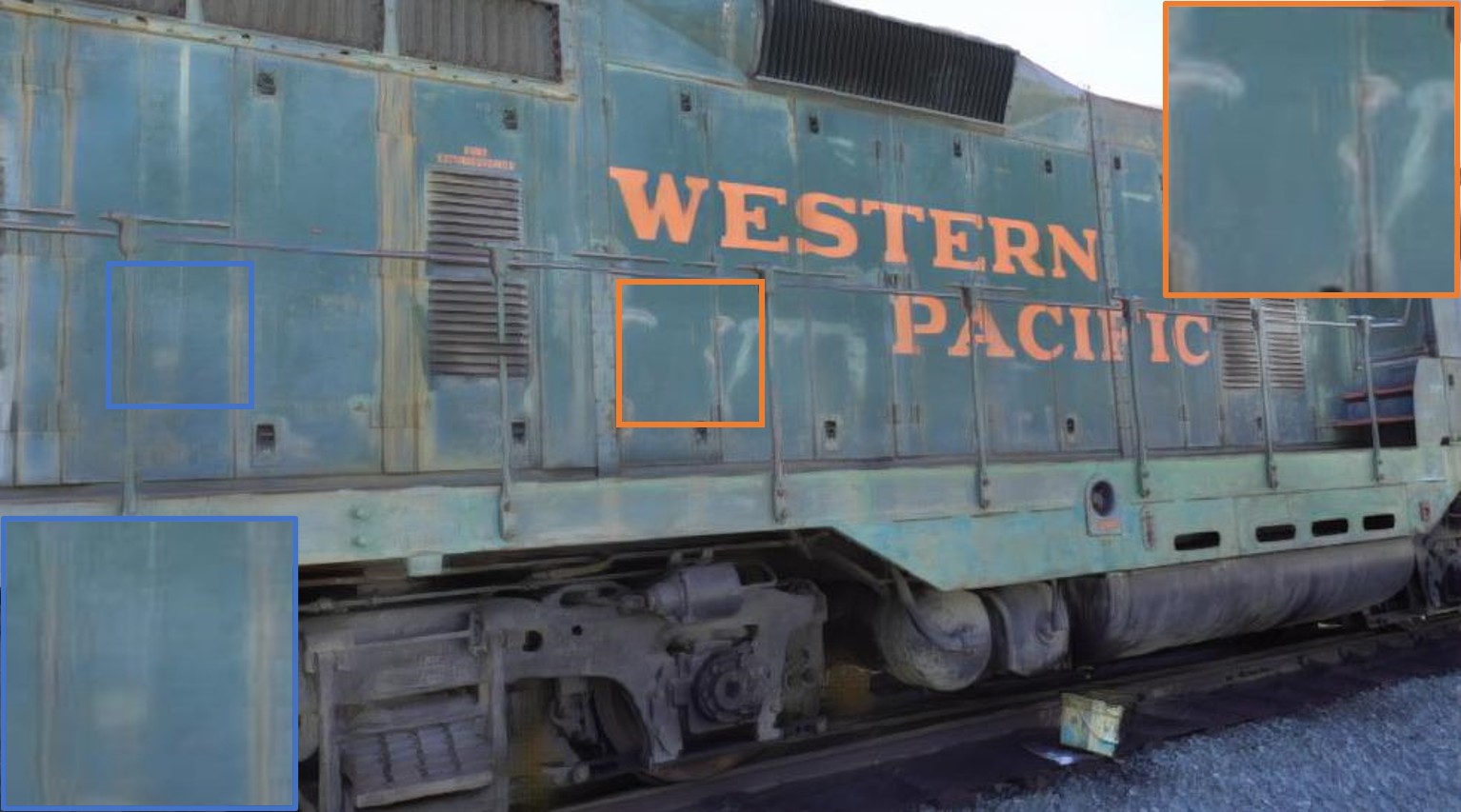} &
\includegraphics[width=0.15\textwidth]{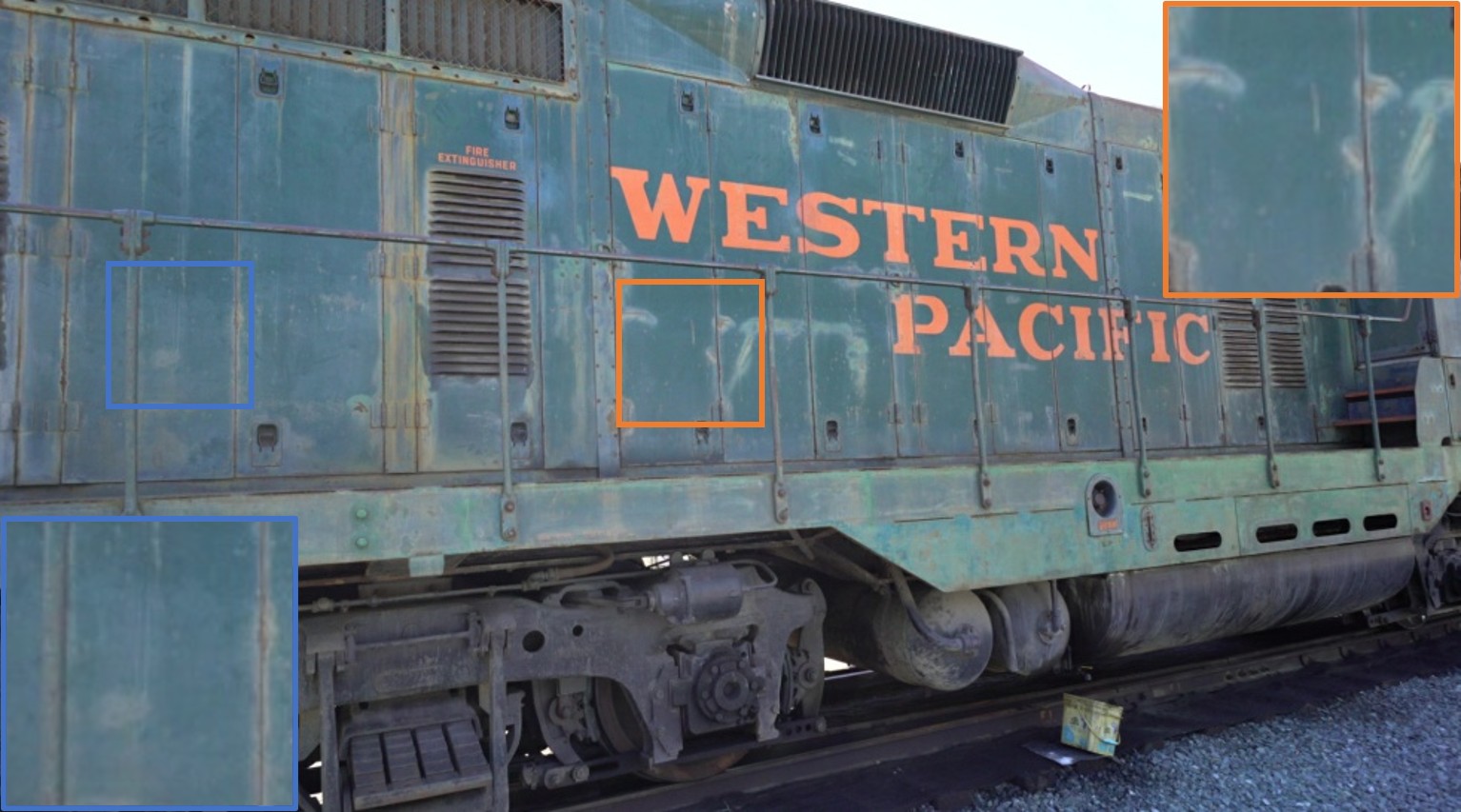} \\
\includegraphics[width=0.15\textwidth]{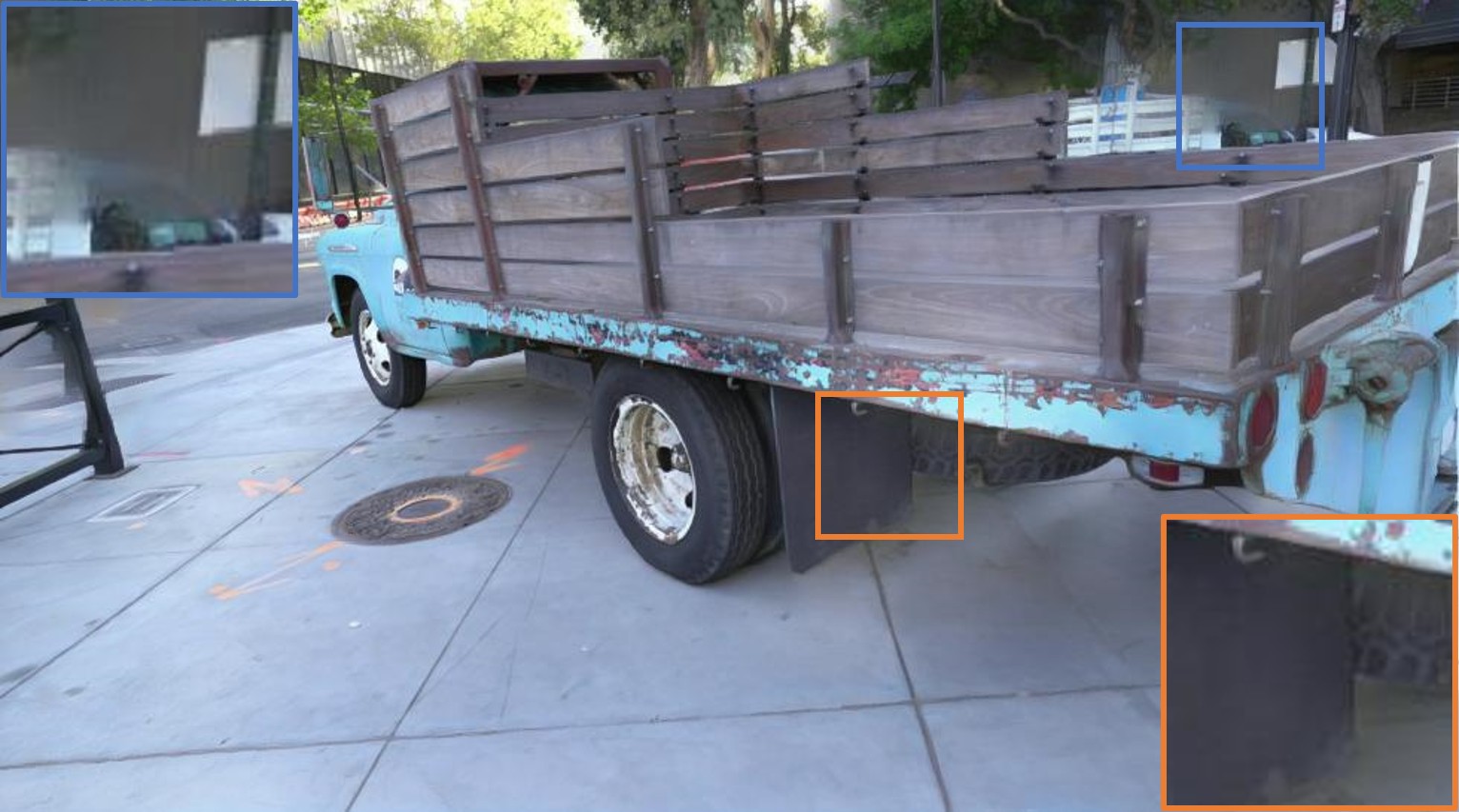} & 
\includegraphics[width=0.15\textwidth]{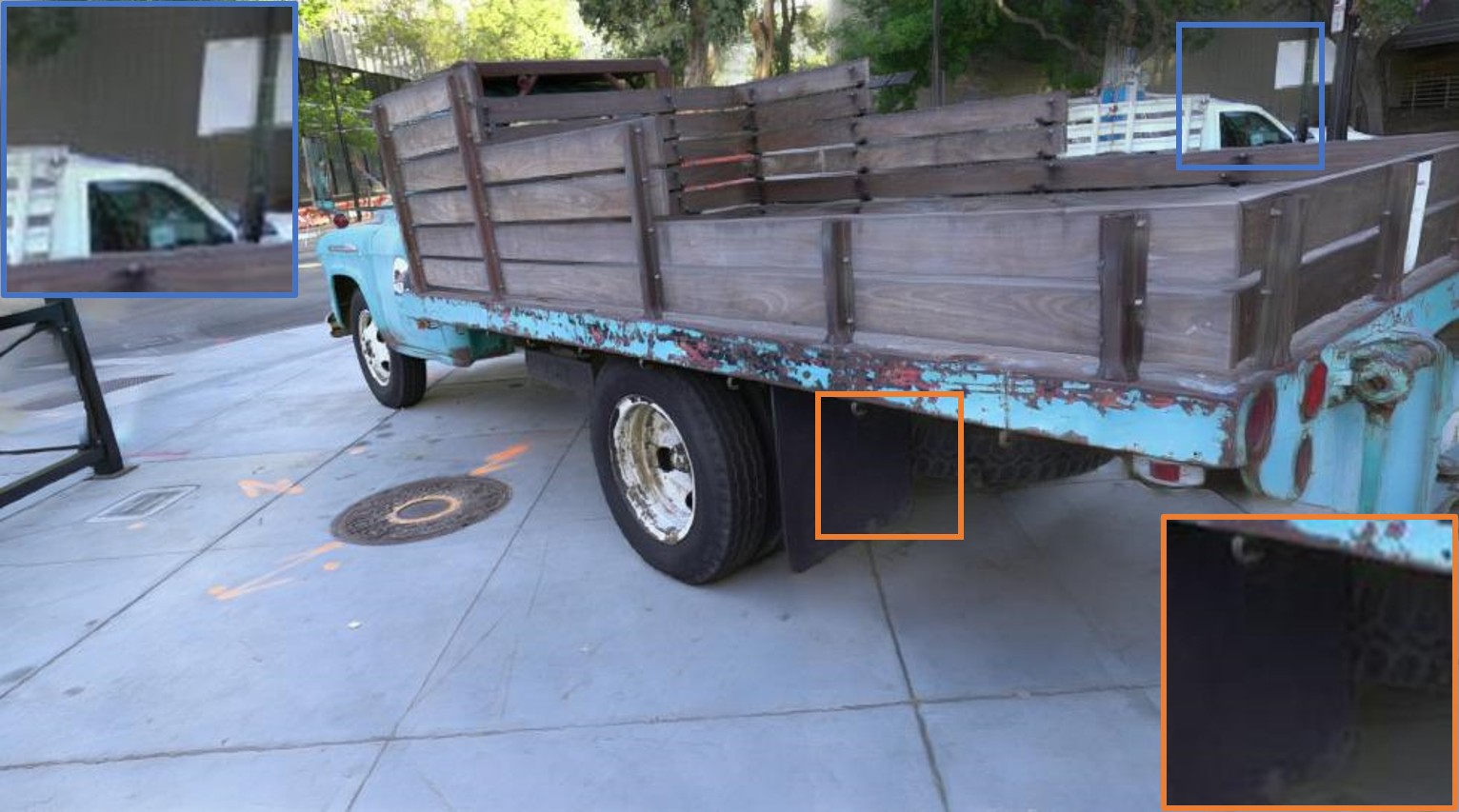} &
\includegraphics[width=0.15\textwidth]{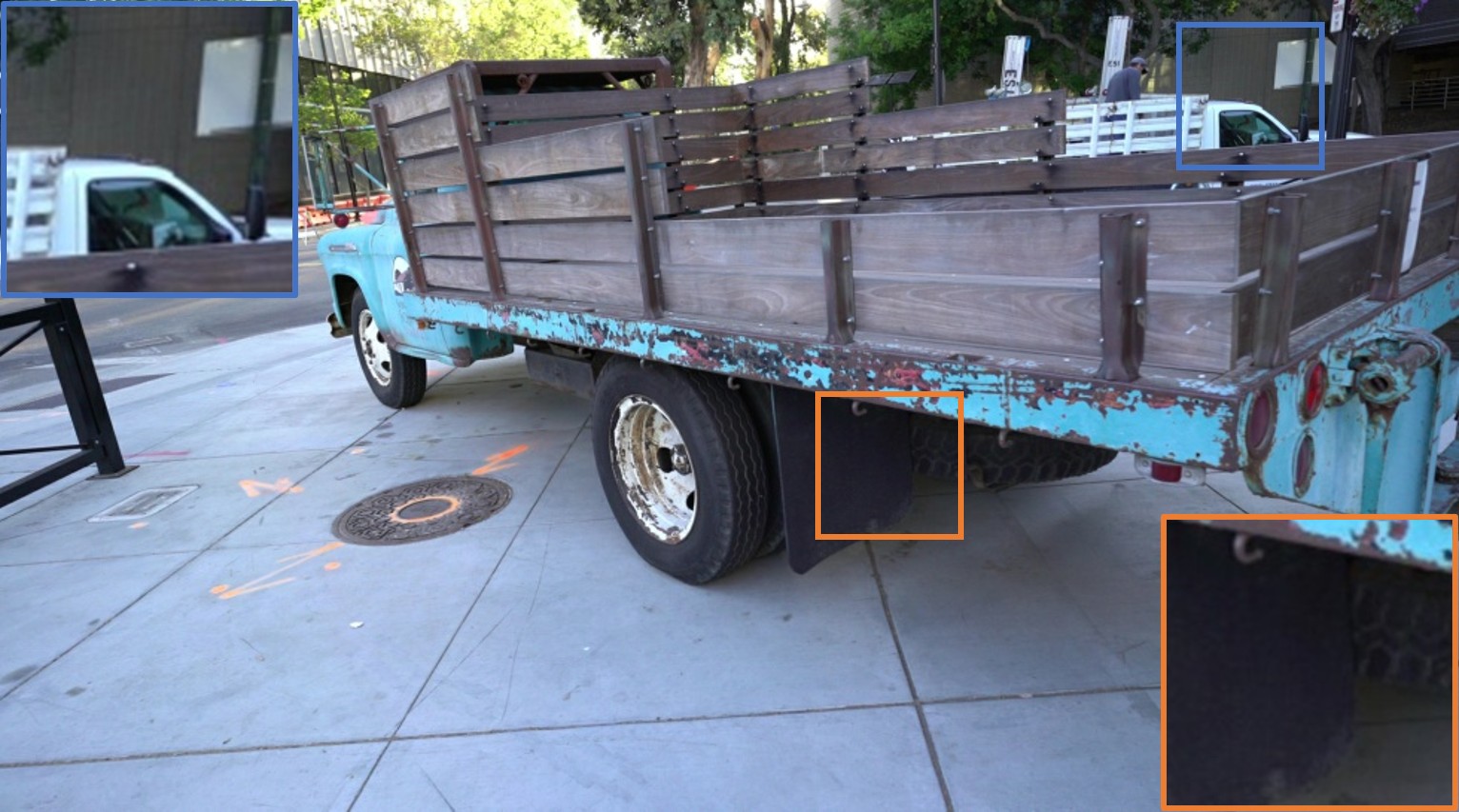} \\
\includegraphics[width=0.15\textwidth]{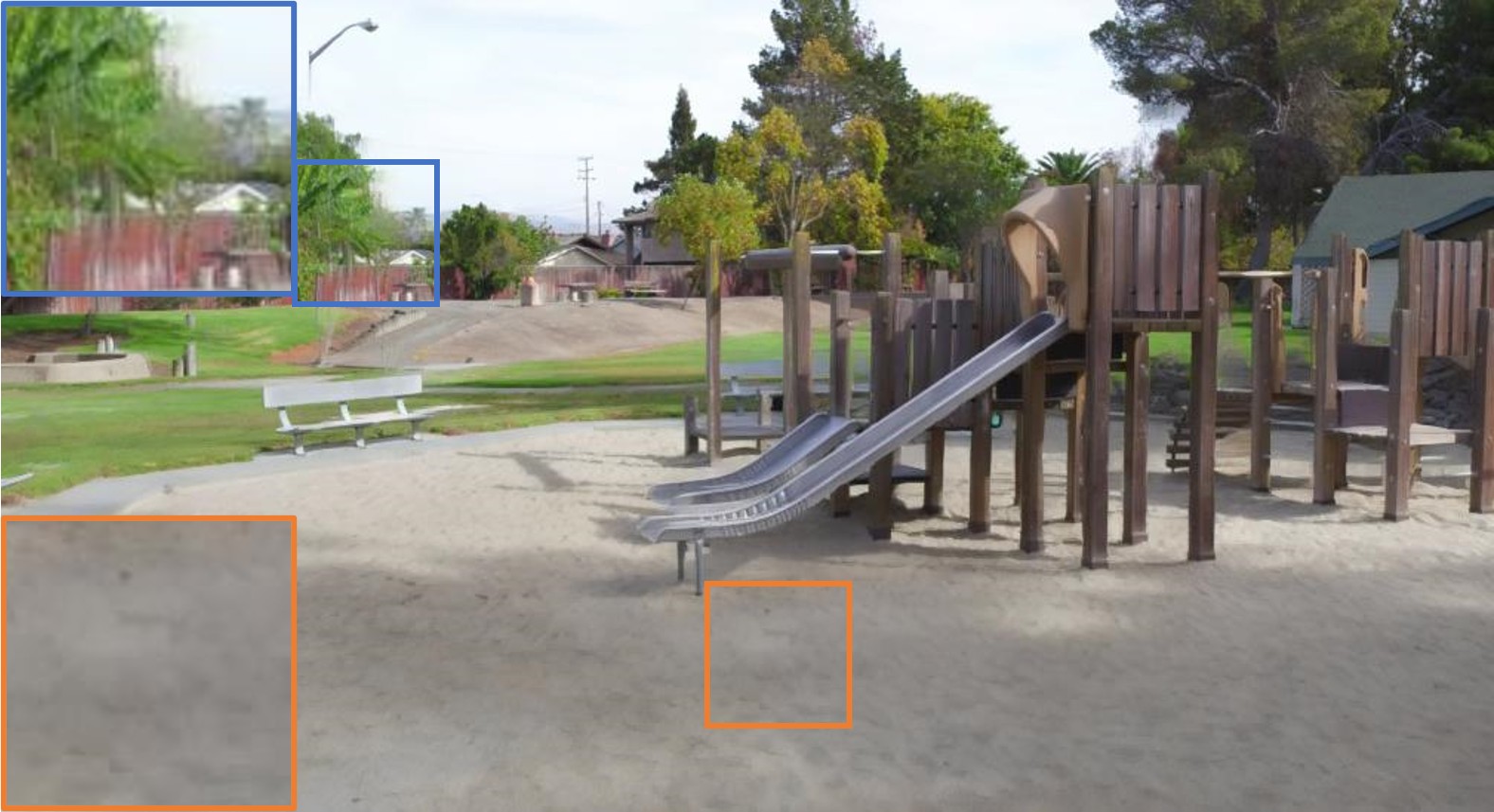} & 
\includegraphics[width=0.15\textwidth]{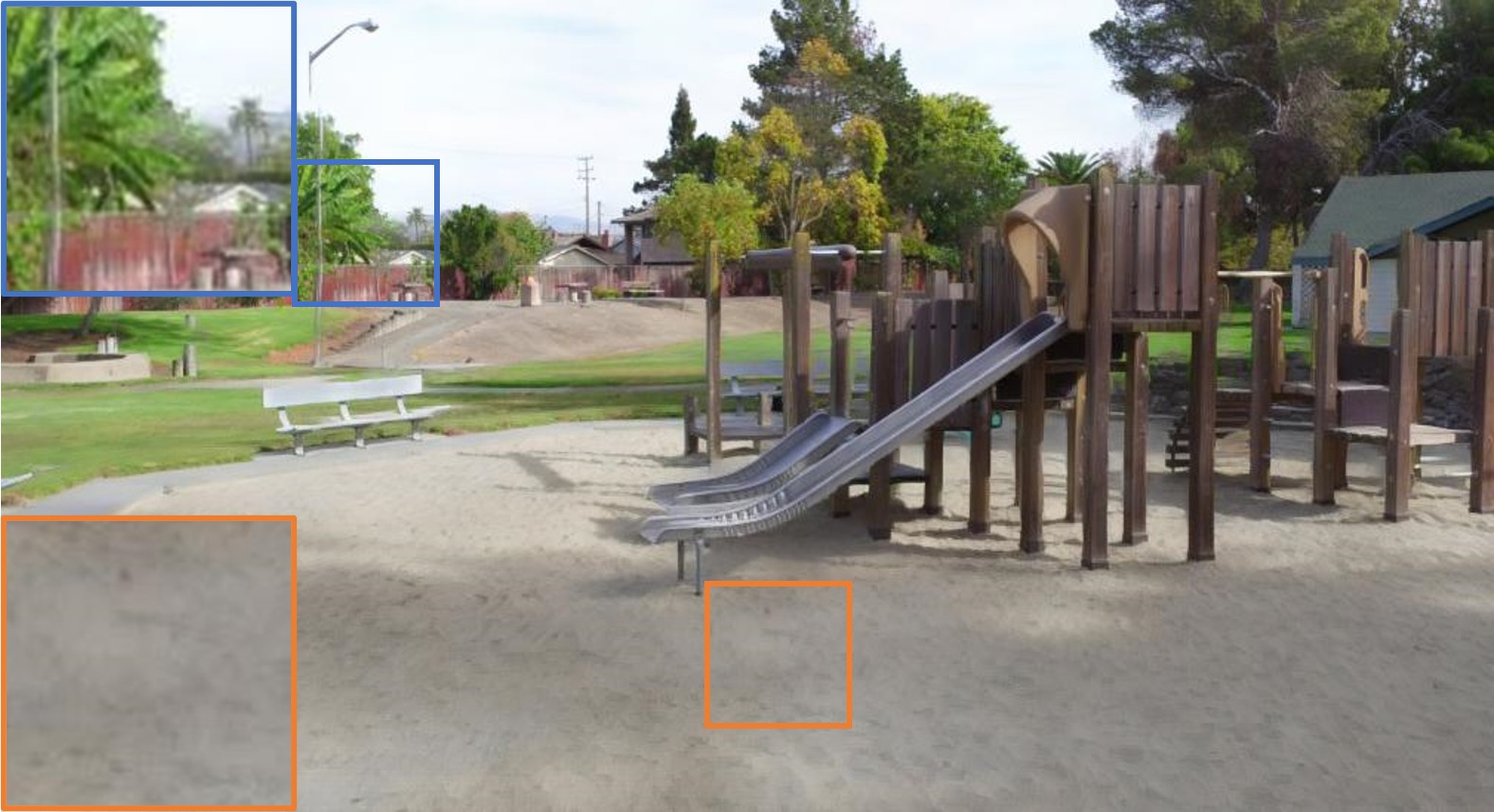} &
\includegraphics[width=0.15\textwidth]{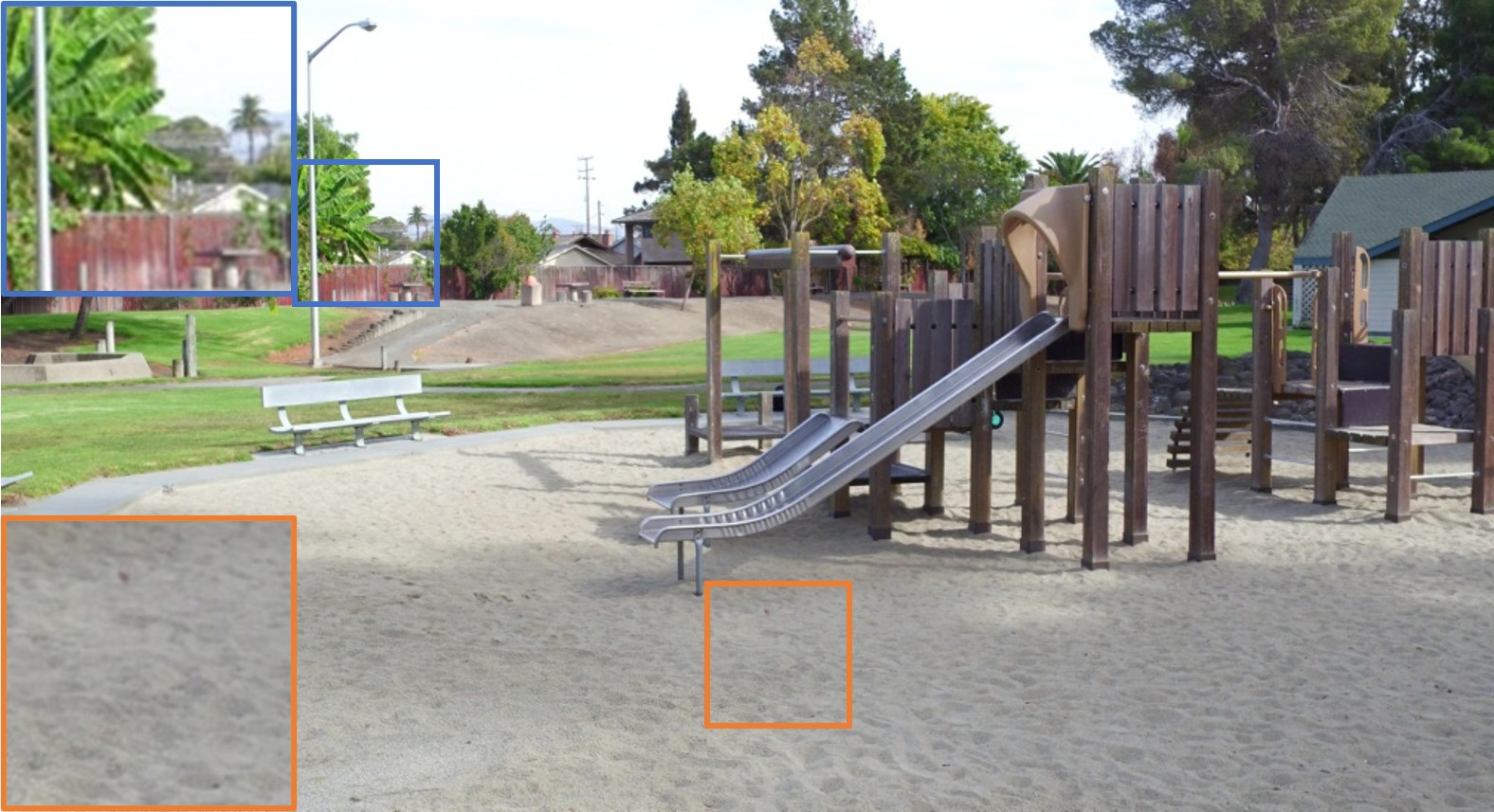} \\
(a) FVS~\cite{Riegler2020FVS} & (b) Ours & (c) Ground Truth \\
\end{tabular}
\setlength{\abovecaptionskip}{2pt} 
\caption{\textbf{Qualitative results of view synthesis.}
}
\vskip -5mm
\label{fig:FVS}
\end{figure}

%% file: figs/fig_visual_comparison.tex
\begin{figure*}[t]
\centering
\footnotesize
\renewcommand{\tabcolsep}{2pt} %
\renewcommand{\arraystretch}{1} %
\begin{tabular}{cccc}

\includegraphics[width=0.24\textwidth]{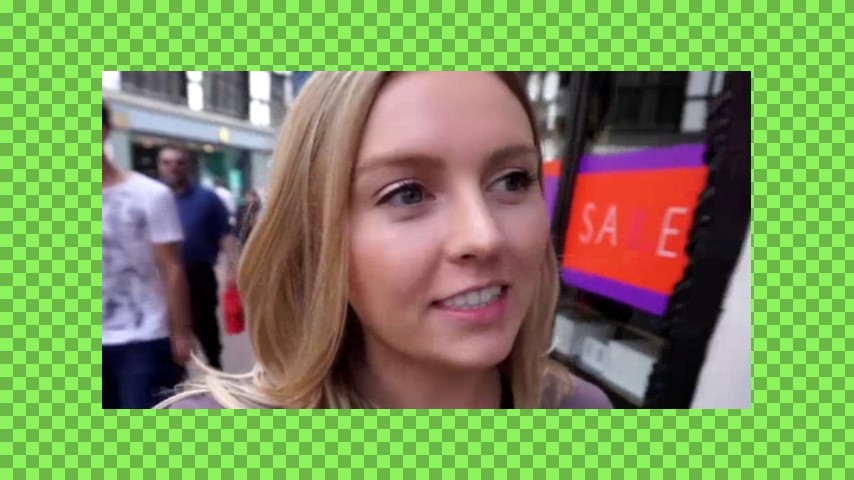} & 
\includegraphics[width=0.24\textwidth]{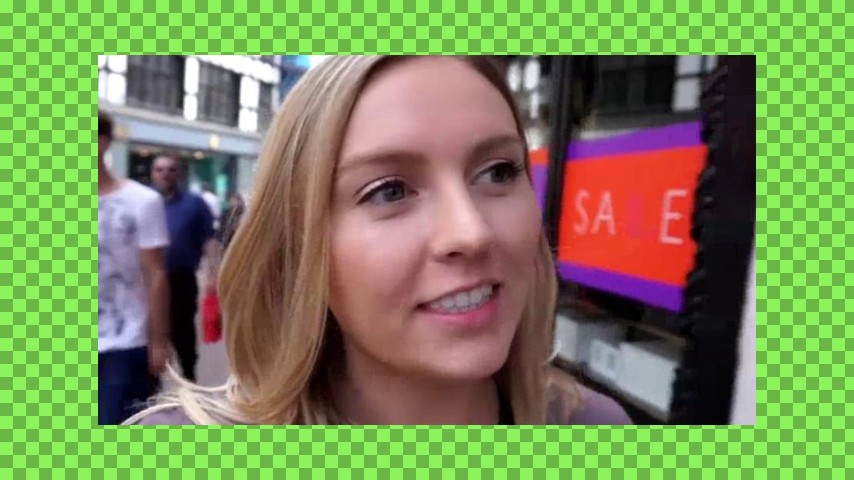} & 
\includegraphics[width=0.24\textwidth]{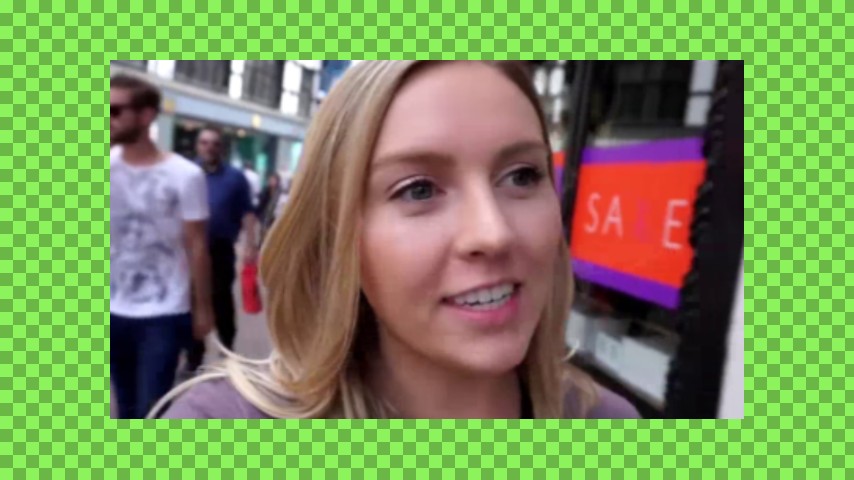} & 
\includegraphics[width=0.24\textwidth]{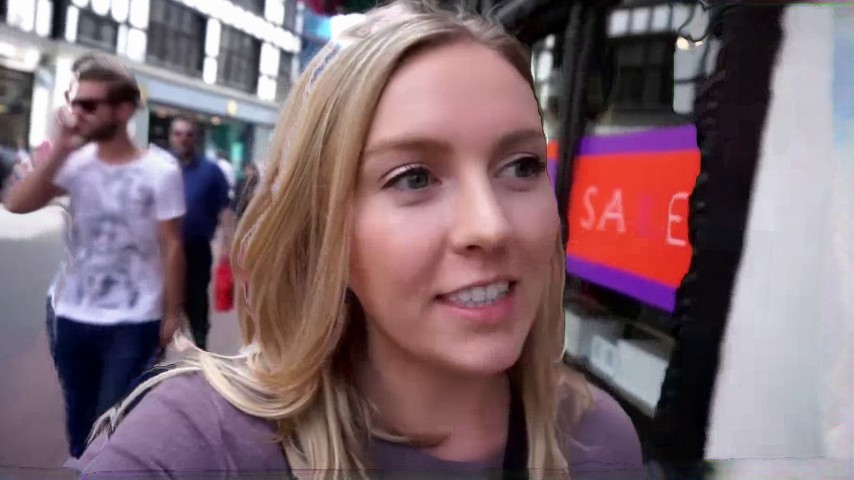} \\
Bundle~\cite{liu2013bundled} & 
L1Stabilizer~\cite{grundmann2011auto} & 
StabNet~\cite{wang2018deep} & DIFRINT~\cite{choi2020deep} \\
\includegraphics[width=0.24\textwidth]{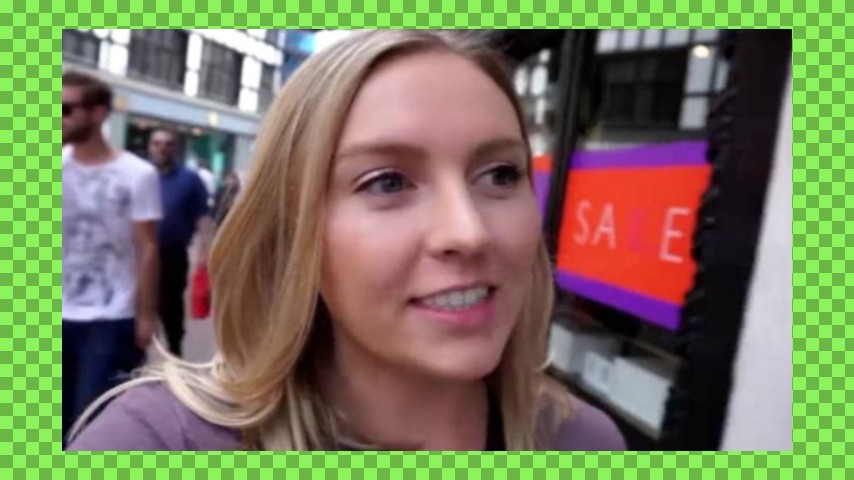} & 
\includegraphics[width=0.24\textwidth]{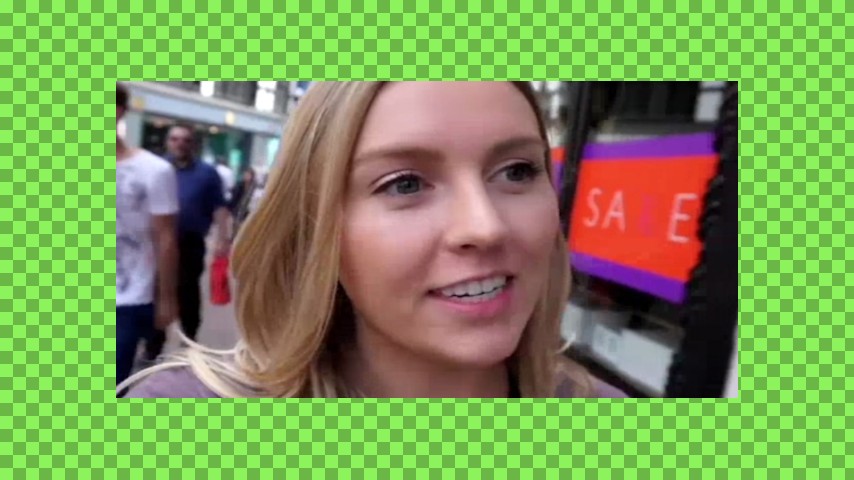} & 
\includegraphics[width=0.24\textwidth]{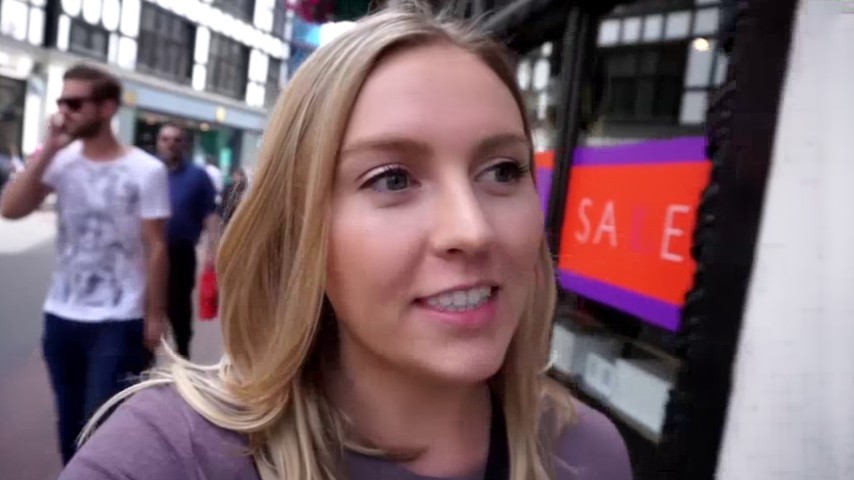} & 
\includegraphics[width=0.24\textwidth]{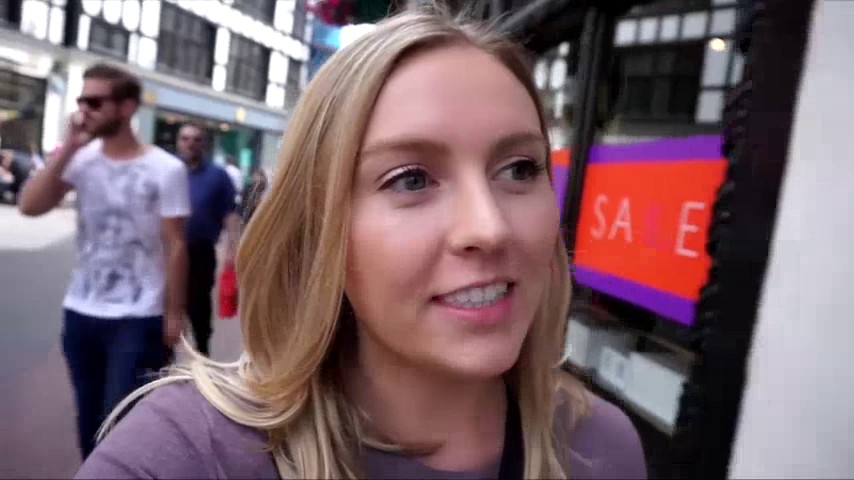} \\
Yu and Ramamoorthi~\cite{yu2020learning} & 
Adobe Premiere Pro 2020 & 
Ours & Input\\
\end{tabular}
\figcapmargin
\caption{\textbf{Visual comparison to state-of-the-art methods.} Our proposed fusion approach does not suffer from aggressive cropping of frame borders and renders stabilized frames with significantly fewer artifacts than DIFRINT~\cite{choi2020deep}. 
We refer the readers to our supplementary material for extensive video comparisons with prior representative methods.
}
\label{fig:visual_comparison}
\end{figure*}

%% file: 7_conclusion.tex
\section{Conclusions}
\label{sec:conclusion}

We have presented a novel method for full-frame video stabilization.
Our core idea is to develop a learning-based fusion approach to aggregate warped contents from multiple neighboring frames in a robust manner.
We explore several design choices, including early/late fusion, heuristic/learned fusion weights, and residual detail transfer, and provide a systematic ablation study to validate the contributions of each component.
Experiments on three public benchmarks demonstrate that our method compares favorably against state-of-the-art video stabilization algorithms.